\documentclass[11pt, a4paper, logo, copyright]{googledeepmind}
\usepackage[authoryear, sort&compress, round]{natbib}

\usepackage{amsmath,amsfonts,bm}

\def\1{\bm{1}}

\DeclareMathAlphabet{\mathsfit}{\encodingdefault}{\sfdefault}{m}{sl}
\SetMathAlphabet{\mathsfit}{bold}{\encodingdefault}{\sfdefault}{bx}{n}

\usepackage{hyperref}
\usepackage{url}

\usepackage{graphicx}
\usepackage{booktabs}
\usepackage{bbm}
\usepackage{lipsum}
\usepackage{enumitem,amssymb,amsmath,amsthm}
\usepackage{multirow}
\usepackage{xspace}
\usepackage{tikz}
\usepackage{dsfont}
\usepackage[english]{babel}
\usepackage{subcaption}
\usepackage{hyperref}
\usepackage{cjhebrew}
\usepackage{CJKutf8}
\usepackage{wrapfig}
\usepackage{caption}
\usepackage{adjustbox}

\usepackage{todonotes}

\let\cite\citep

\newcommand{\note}[5]{}

\newcommand{\eclektic}[0]{\textsc{ECLeKTic}\xspace}

\title{\eclektic: a Novel Challenge Set for Evaluation of Cross-Lingual Knowledge Transfer}

\correspondingauthor{omer.goldman@gmail.com, \{urishaham,matane\}@google.com}

\newcommand{\GR}{$\gamma$}
\newcommand{\BIU}{$\beta$}
\newcommand{\DM}{$\delta$}

\author[\GR \BIU*]{Omer Goldman}
\author[\GR*]{Uri Shaham}
\author[\GR]{Dan Malkin}
\author[\GR]{Sivan Eiger}
\author[\GR]{Avinatan Hassidim}
\author[\GR]{Yossi Matias}
\author[\DM]{Joshua Maynez}
\author[\GR]{Adi Mayrav Gilady}
\author[\DM]{Jason Riesa}
\author[\DM]{Shruti Rijhwani}
\author[\DM]{Laura Rimell}
\author[\GR]{Idan Szpektor}
\author[\GR]{Reut Tsarfaty}
\author[\GR]{Matan Eyal}

\affil[*]{Equal Contribution}
\affil[\GR]{Google Research}
\affil[\BIU]{Bar-Ilan University}
\affil[\DM]{Google DeepMind}

\begin{abstract}
To achieve equitable performance across languages, large language models (LLMs) must be able to abstract knowledge beyond the language in which it was learnt. However, the current literature lacks reliable ways to measure LLMs' capability of such cross-lingual knowledge transfer.
To that end, we present \mbox{\eclektic},\footnote{The dataset is available at \url{https://www.kaggle.com/datasets/googleai/eclektic}
} a multilingual closed-book QA dataset that Evaluates Cross-Lingual Knowledge Transfer in a simple, black-box manner.
Concretely, we used the presence and absence of Wikipedia articles in 12 languages to 
detect pieces of information that were likely available during pre-training in  one of the languages but not in the others. We   curate \mbox{\eclektic} as a set of fact-seeking questions over this kind of information, in all the different languages. 
Therefore, in order to solve \mbox{\eclektic} the model is required to transfer knowledge between languages.
We evaluated 8 LLMs and showed that current SOTA models struggle to effectively share knowledge across languages, even if they can predict the answer for questions in the language in which the knowledge was acquired. 
\end{abstract}

\begin{document}

\maketitle

\footnotetext{The dataset is available at \url{https://www.kaggle.com/datasets/googleai/eclektic}}

\section{Introduction}

Multilingual large language models (LLMs; \citealp{gemini, llama, openai2024gpt4}, inter alia) should perform similarly and consistently in all languages they were trained on, and crucially, possess an equivalent depth of knowledge across these languages. 
This would make models more similar to humans that decouple knowledge from language. %
But more importantly, it would make LLMs more efficient %
and provide equitable access to a wider range of the world's knowledge to speakers of all languages. 

However, measuring knowledge-sharing across  languages is nontrivial. 
When a model provides consistent answers to the same question across different languages, it's difficult to determine the underlying cause. 
Does this consistency stem from a single, language-independent representation of knowledge -- or did the model simply learn from its training data the same fact   in each language separately? %
Due to this ambiguity, previous works focused on cross-lingual \textit{consistency} alone, measuring whether models answer the same across languages, %
without investigating the reason for that consistency \citep{jiang-etal-2020-x, kassner-etal-2021-multilingual, ohmer-etal-2023-separating, qi-etal-2023-cross}. 
These works translated entire datasets, inherently testing 
facts that the model was likely exposed to in multiple languages during training, thus making it impossible to distinguish true knowledge sharing from parallel exposure.%

In order to determine whether cross-lingual consistency in LLMs is incidental or principled we 
present \eclektic, to \textbf{E}valuate \textbf{C}ross-\textbf{L}ingual \textbf{K}nowledge \textbf{T}ransfer in a simple, black-box manner. We achieve that by building a closed-book Question Answering (QA) evaluation set that carefully targets specific facts that were likely acquired only in one language. Consider, for example, \hfill a \hfill question \hfill asking 

\begin{wrapfigure}{r}{0.5\textwidth}
  \vspace{-20pt}
  \begin{center}
    \includegraphics[width=0.5\columnwidth]{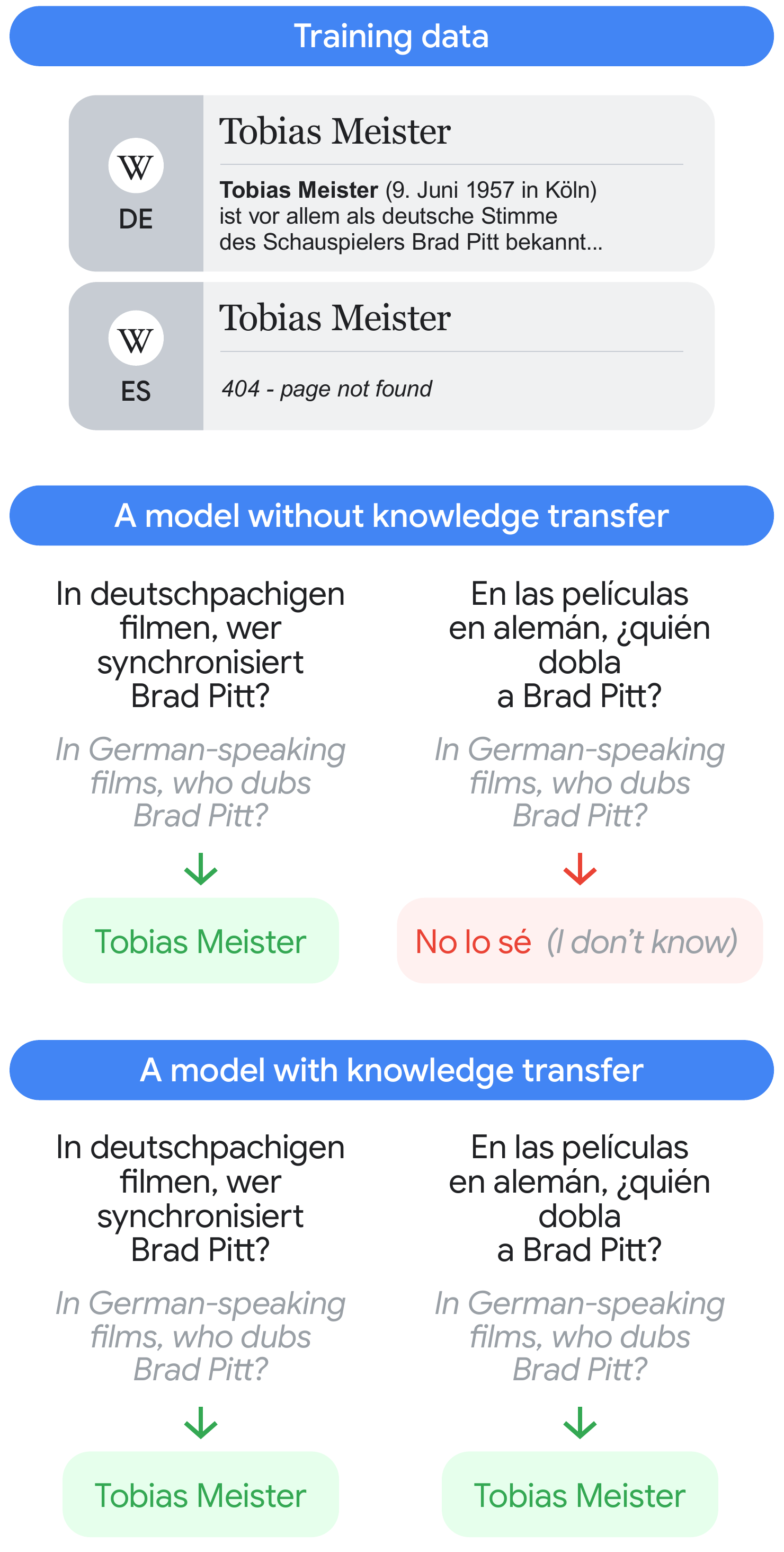}
  \end{center}
    \caption{A model incapable of cross-lingual knowledge transfer (middle box) can only answer factual questions in their \textcolor{blue}{\textbf{source language}}, that is, the language in which the information appeared in its training. It cannot answer the same question when translated into another \textcolor{orange}{\textbf{target language}}. A transfer capable model (bottom box), is able to answer questions no matter the language. \mbox{\textsc{ECLeKTic}} allows distinguishing between the two by targeting facts that unevenly distributed in the model's training data.}%
  \vspace{-10pt}
\label{fig:1st_page}
\end{wrapfigure}

\noindent who usually dubs characters played by Brad Pitt in German movies. Since in the German part of the internet it is well-known that the answer is Tobias Meister, a modern LLM answers this question easily when asked in German. However, to answer that question in other languages, in which the internet contains little or no evidence of the underlying fact, LLMs that are not able to internally retrieve and share across languages that German knowledge may struggle (see \autoref{fig:1st_page}).

In this paper we present \eclektic to evaluate cross-lingual transfer by targeting such language-specific facts. 
We identified those facts using 
Wikipedia articles in 12 languages that lack equivalents in the other 11, such as the article on the aforementioned Tobias Meister, that exists only on the German Wikipedia. We annotated each such article with a fact-seeking question and answer, and translated them to all other tested languages in our benchmark. The entire generation and translation process was done by human annotators, based on LLM-generated suggestions. The result is a set of 4,224 question/answer pairs, each addressing a fact primarily known in one language (henceforth, the \textit{source language}) but written in one of the other 11 language (the \textit{target languages}). Additionally, each question-answer pair is accompanied by the Wikipedia paragraphs on which it was based, in its original (source) language.

We evaluated LLMs on our benchmark using a judge model \cite{chiang-lee-2023-large, zheng2023judging}, and aggregate example-level correctness 
into two metrics: %
\textit{overall success} that reflects the ability to solve \eclektic and answer questions in their source and target languages, and \textit{transfer ability} that only measures the ability to transfer \textit{correct} answers.

We experimented with 8 top-performing models, open-source and proprietary, and our results demonstrate that \textsc{ECLeKTic} poses a significant challenge across the board. The best performing model, Gemini 2.0 Pro, achieves \textit{overall success} of 41.6\% and manages to transfer only 65.0\% of the facts it was able to retrieve in the respective source language. 
Breaking down the results by source and target language, we show that shared script is a major factor in the ease of transfer, corroborating findings from previous works \cite{malkin-etal-2022-balanced,qi-etal-2023-cross,ifergan2024beneath}. %
Moreover, we showed, by examining models of different sizes from the Qwen 2.5 model series \cite{qwen25}, that that bigger models are not able to transfer more knowledge in relative terms, i.e., in terms of \textit{transfer ability}, although they are more successful in terms of \textit{overall success}.

All in all, this work contributes a novel benchmark for cross-lingual transfer evaluation, along with its construction, metric definition, and evaluation process. \eclektic shows  
that there is a considerable headroom for further research into more capable and consistent multilingual models.

\begin{figure}[t]
    \centering
    \includegraphics[width=\textwidth]{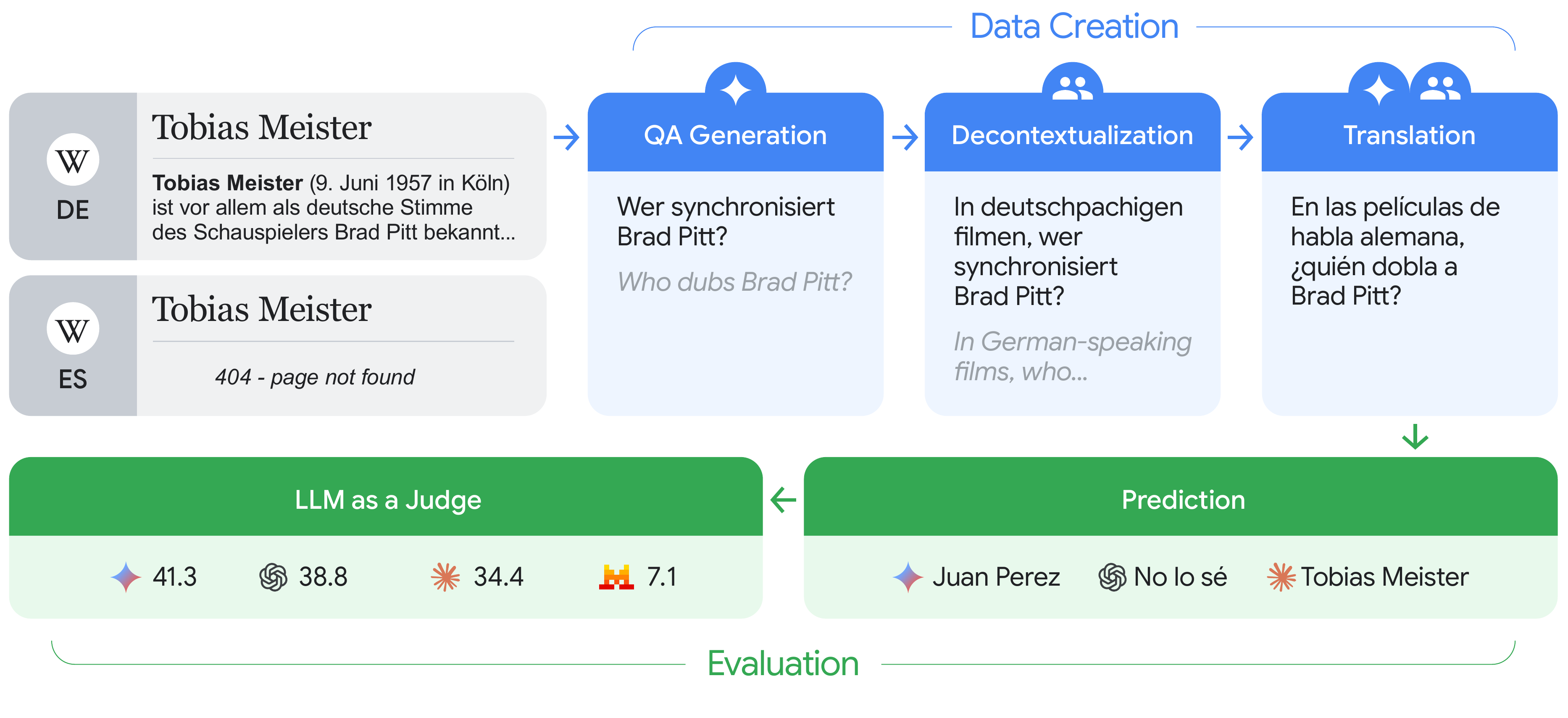}
    \caption{A schematic overview of the creation of \eclektic and its application in evaluating language models. %
    }
    \label{fig:pipeline}
\end{figure}

\section{Constructing \eclektic}
\label{sec:data_creation}

\eclektic is a QA benchmark that includes questions of which answers were exposed to the model in a \textit{single} language during pre-training. 
As a basis for such question/answer pairs, we selected articles in Wikipedia that only exist in one of the 12 target languages, and not in any of the others (English, French, German, Hebrew, Hindi, Indonesian, Italian, Japanese, Korean, Mandarin Chinese, Portuguese, and Spanish). Concretely, we analyzed the July 2023 dump of Wikipedia and for each language sampled 100 articles that contain at least 200 characters, had at least 100 views during 2023,\footnote{Statistics available on: \url{https://stats.wikimedia.org/##/all-wikipedia-projects}.} and most importantly do not have equivalent articles in any of the other 11 languages. 

The annotation process was done in two stages. In the first, annotators were presented with Gemini-suggested a 
question-and-answer pair based on the 10 first sentences of each Wikipedia page. The annotators verified that the questions are based on the context, answerable from the context, and that the question is particularly relevant to the language in question, e.g., it does not relate to general common knowledge or scientific  facts. The annotators also checked whether the question is answerable in a closed book setting, i.e., it does not refer explicitly to the context or mention the answer. %
Questions and answers that did not meet these criteria were discarded. 
Lastly, the annotators made sure that questions contain all the information needed to be answerable when translated. For example, a question in Hebrew relating to ``the supreme court'' was disambiguated to explicitly mention ``the Israeli supreme court''. Named entities were also clarified similarly, so a question referring to ``Ambev'' was modified to refer to ``the Brazilian brewing company, Ambev''.

In the second stage, the questions, answers and their contexts were automatically translated to the other 11 languages by Gemini. Then they were given to another set of translation-expert annotators for verification and modification. Specifically, the annotators corrected the translation of uncommon named entities and verified that appropriate transliteration rules were applied. At this stage some examples were also discarded if they proved to be untranslatable. For example, when a question explicitly refers to the meaning of a word in the source language. To overcome the difficulties in 
translation between non-trivial language pairs, this stage was done through English as a pivot language. 
The annotators of this stage were also tasked with verifying the work of the previous annotators and discard any questions that do not comply with any of the aforementioned criteria. All in all, the human annotation effort amounted to 400 cumulative hours, performed by a team of 27 annotators.

The complete annotation process is depicted in \autoref{fig:pipeline}. 
The prompts that were given to the model and the annotation guidelines that were given to the annotators in the data creation process are detailed in \hyperlink{sec:guidelines}{Appendices A} and \ref{sec:prompts}.

\subsection{Assumptions in \eclektic} 
\label{sec:assumptions}

There are several important assumptions we made in the construction of \eclektic, which we make explicit here.

First, we assume that all the tested models are exposed to the articles we extracted from Wikipedia. Moreover, we assume that models were exposed to the information in those articles in their respective source languages multiple times during their pretraining. %
This assumption is straightforward given that in practice Wikipedia itself is repeated in the pre-training data of most LLMs due to the quality of its texts \cite{gpt3, palm}. Moreover, the existence of a Wikipedia article is likely to reflect a general interest of online speakers in the same topic so the information is also likely to be repeated in the same language outside of Wikipedia. This is even more straightforward given that we only targeted articles with significant yearly view count. 
Conversely, we assume that the absence of an article from a certain Wikipedia reflects the lack of interest of online speakers in that topic. The information on that topic is therefore assumed to appear sparsely on the internet, if at all. %

Importantly, we assume a link between the abundance of information in a model's training data and the accessibility of that information to the model from its weights. That is, we assume that \textit{without means of cross-lingual transfer} it will be easier for a model to retrieve information in a certain language when it has seen this information more frequently in that language during pretraining.\footnote{See \citet{verma2024van} for a verification of a similar hypothesis regarding frequency and accessibility in computer vision.}

Together, these assumptions allow us to treat the (non)existence of a fact in Wikipedia as an approximate to the 
(in)accessibility of that fact to a model trained on online data. In other words, we can treat questions about topics missing in Wikipedia in a certain language, as questions that require cross-lingual transfer to be answered.

\subsection{The \eclektic Benchmark}

\begin{figure}[t]
\begin{minipage}[t]{0.49\textwidth}
    \centering
    \includegraphics[width=\linewidth]{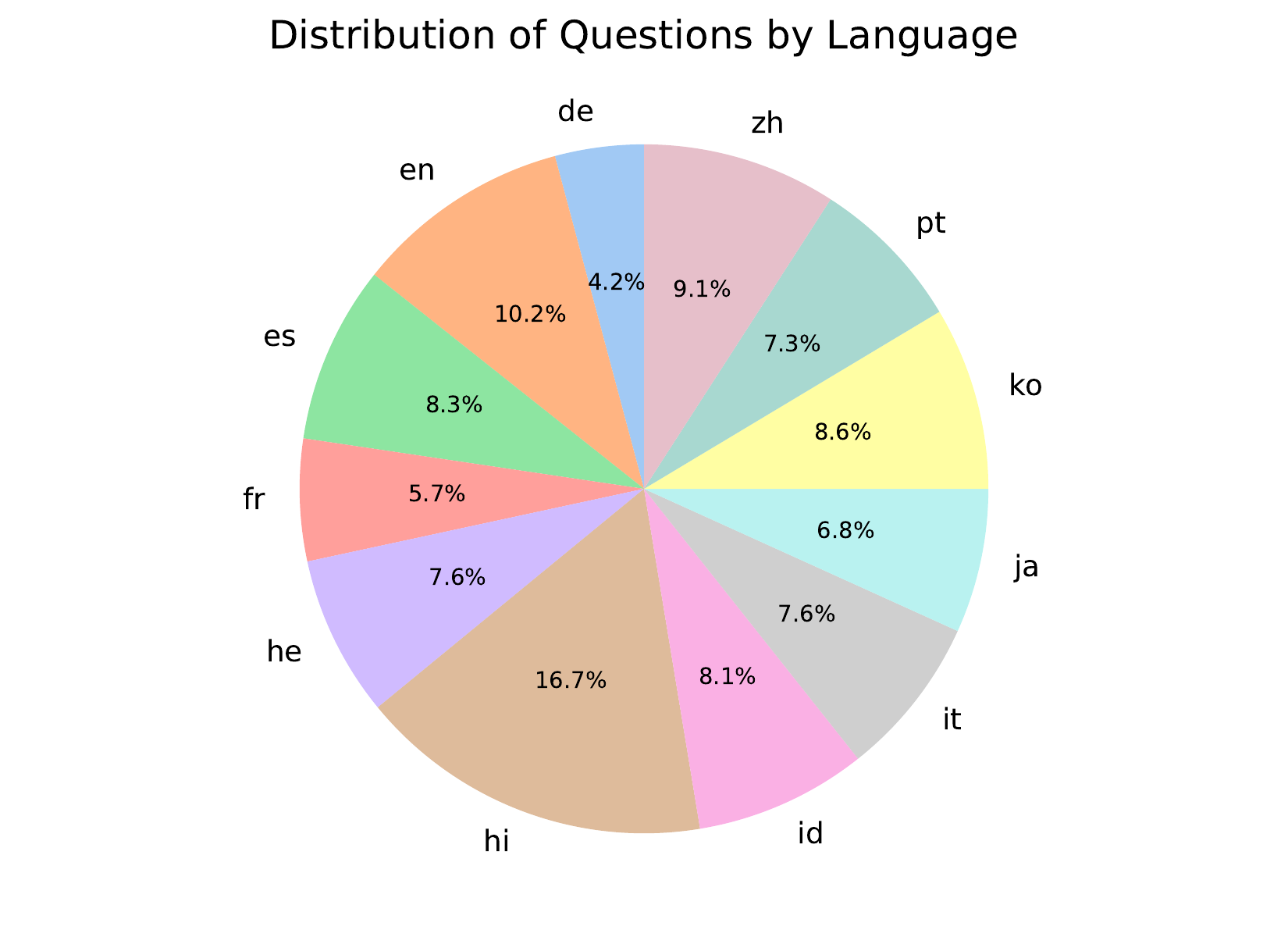}
    \captionof{figure}{Break down of the examples in \eclektic by source language.}
    \label{fig:lang-breakdown}
\end{minipage}\hfill
\begin{minipage}[t]{0.49\textwidth}
    \centering
    \includegraphics[width=\linewidth]{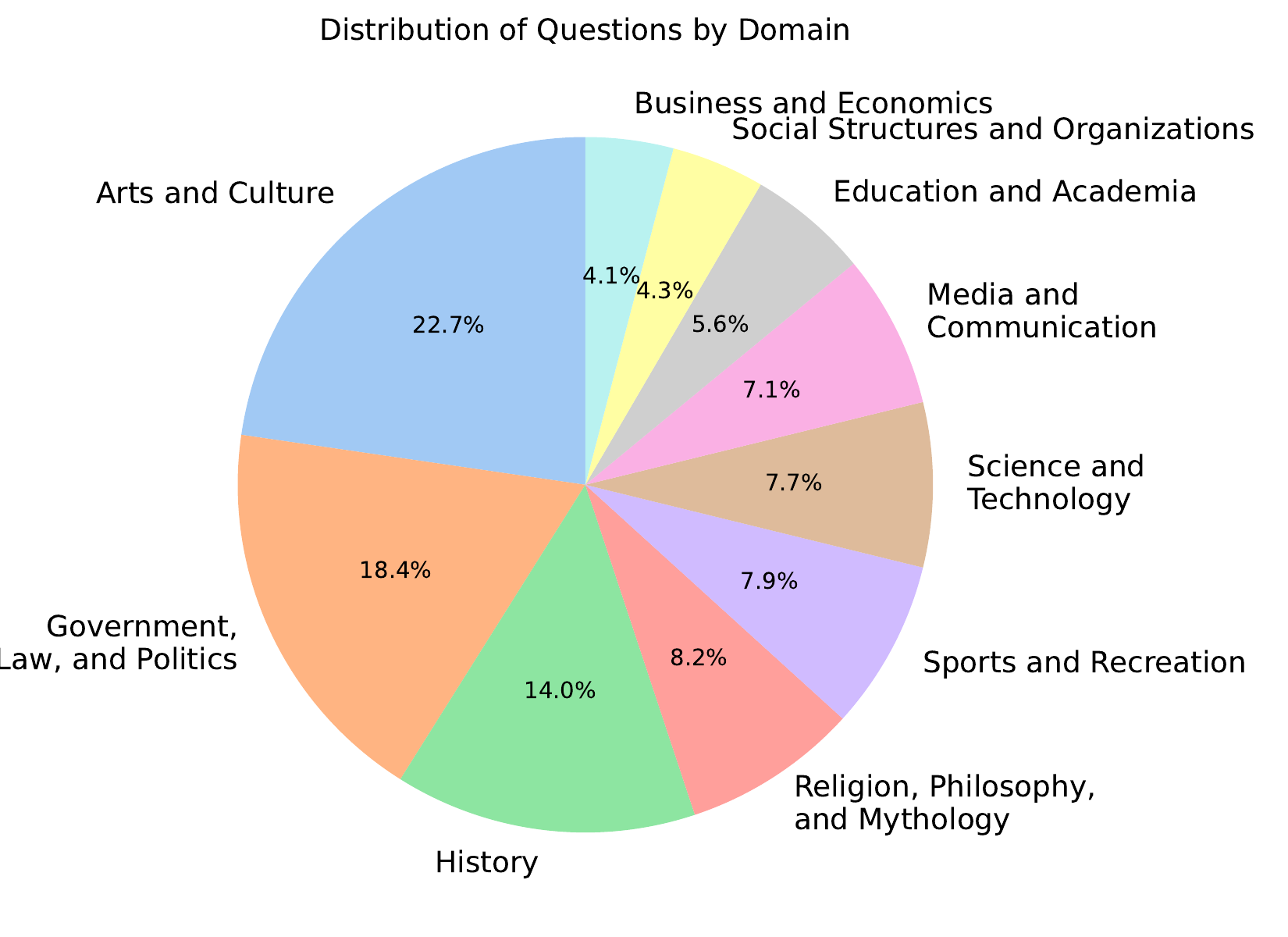}
    \captionof{figure}{Break down of the examples in \eclektic by domains of knowledge. A breakdown done separately per source language can be found in \autoref{sec:per-language-domains}.}
    \label{fig:domain-breakdown}
\end{minipage}
\end{figure}

The resulting benchmark includes 4608 question/answer pairs in 12 languages. 4224 of which are evaluated as target examples, i.e., examples in languages where the fact is not available to the model, 
and 384 are questions in their source language, used to evaluate the amount of information to be transferred. The abundance of each source language in \eclektic is in \autoref{fig:lang-breakdown}.

\autoref{fig:domain-breakdown} details the distribution of questions across 10 domains detected automatically by Gemini.\footnote{We verified the domain classification by sampling 100 questions. All of them were classified correctly by Gemini.} It reflects the various topics covered by \eclektic and verifies that our method of construction did not result in homogeneous questions. When breaking down the questions in \eclektic by domain separately for each source language, it is possible to see the we indeed captured the unique interests of speakers of each language. For example, "religion, philosophy, and mythology" is over-represented in the examples originated in Hebrew and Hindi, in accordance to the unique distribution of religions among their speakers, "arts and culture" is over represented in French and Italian, and "history" -- in Chinese. The domain breakdown per-language is detailed in \autoref{sec:per-language-domains}.

\section{\eclektic Metrics}

To empirically assess the cross-lingual knowledge-transfer capabilities over \eclektic, we devise two 
metrics that accompany the benchmark. The first is \textit{overall success}, which measures the extent to which a model succeeds in answering correctly the questions in \eclektic, in both source and target language. The second is \textit{transfer score}, which measures the success in the knowledge transfer itself,
taking into account 
only  the questions answered correctly in their source language.

Formally, for each question/answer pair we define \textit{example-level success}, based on the target language in which they are written $l_t$ and on the source language from which they were translated $l_s$, as
$$S^{l_t}_{q,a} = \mathds{1}\big(M(q_{l_t})=a_{l_t}\big) \land \mathds{1}\big(M(q_{l_s})=a_{l_s}\big)$$
\noindent where $M(q_l)$ is the prediction of model $M$ for question $q$ in language $l$. $S^{l_t}_{q,a}$ is a binary score that indicates 
whether a question is correctly answered in its source language $l_s$ as well as in another target language $l_t$, capturing positive knowledge sharing between the two languages. 
Given a set of examples $D$, the final metric \emph{overall} simply averages across all question/answer pairs.
$$S^{\text{overall}} = \frac{\sum_{(\{q,a\} \in D, l_t \in L)}{S^{l_t}_{q,a}}}{|D|}$$
$S^{\text{overall}}$ is the main metric of \textsc{ECLeKTic}.

To strictly measure the knowledge transfer probability, we consider only question/answer pairs for which the model provided the correct answer in the source language: $K=\{(q,a) | M(q_{l_s})=a_{l_s}\}$.
We then define \textit{transfer score} as the number of questions that were answered correctly by the model in $l_t$ given that they were answered correctly in $l_s$: 
$$S^{\text{transfer}} = \frac{\sum_{(\{q,a\} \in K, l_t \in L)}{S^{l_t}_{q,a}}}{|K|}$$
Note that this metric does not reflect the number of questions answered correctly in their source language.  As a result, it can be maximized even by weak models, for which $|K|$ is small, as long as they answer the same questions in all languages.

To determine whether a model gives the correct answer to a specific question in a specific language $M(q_{l})\stackrel{?}{=}a_{l}$ we use an LLM as a judge \cite{zheng2023judging}. Specifically, we let Gemini 2.0 Flash verify the answer's correctness based on the question and the translated context \cite{zhou2025graders}.\footnote{The prompt used for judgement is in \autoref{sec:prompts}.} 

\subsection{String Matching Evaluation}

Evaluation by an LLM tends to be better than automatic metrics that are less correlated with human judgements \cite{chen-etal-2019-evaluating}. However, we introduce an automatic string matching metric to accompany the model-based metric.\footnote{The code for string-based evaluation of predictions can be found at \url{https://github.com/omagolda/ECLeKTic-string-metrics}} This is done due to its ease of execution and to verify that the judge model does not introduce biases of its own.

In this metric we define the correctness of answer as word-level recall $R(p,g)$, i.e., the portion of words in the gold answer $g$ that are in the model's prediction $p$.\footnote{In Chinese and Japanese we counted every Chinese character as a word, for simplicity's sake.} Subsequently, the example-level success is then defined as 
$$S^{l_t}_{q,a} = R\big(M(q_{l_t}),a_{l_t}\big) \cdot R\big(M(q_{l_s}),a_{l_s}\big).$$
$S^{\text{overall}}$ then remain the averaged example-level success over all examples. But as we can not say with this metric which examples are in $K$, we amend $S^{\text{transfer}}$ to be a weighted average over the entire benchmark $D$ where the recall over the source examples serve as weights
$$S^{\text{transfer}} = \frac{\sum_{(\{q,a\} \in D, l_t \in L)}{S^{l_t}_{q,a}}}{ \sum_{(\{q,a\} \in D)} R(M(q_{l_s}),a_{l_s}) }$$

\section{Experiments}

We used \textsc{ECLeKTic} to evaluate cross-lingual knowledge transfer in several leading models. We included \hfill open \hfill models, \hfill the \hfill latest \hfill instruction-tuned \hfill versions \hfill of \hfill Gemma,\footnote{\url{https://huggingface.co/google/gemma-2-9b-it}} \hfill Mistral,\footnote{\url{https://huggingface.co/mistralai/Mistral-Nemo-Instruct-2407}} \hfill Qwen\footnote{\url{https://huggingface.co/Qwen/Qwen2.5-7B-Instruct}}, \hfill and

\begin{wrapfigure}{r}{0.5\textwidth}
  \vspace{-20pt}
  \begin{center}
    \includegraphics[width=0.5\columnwidth]{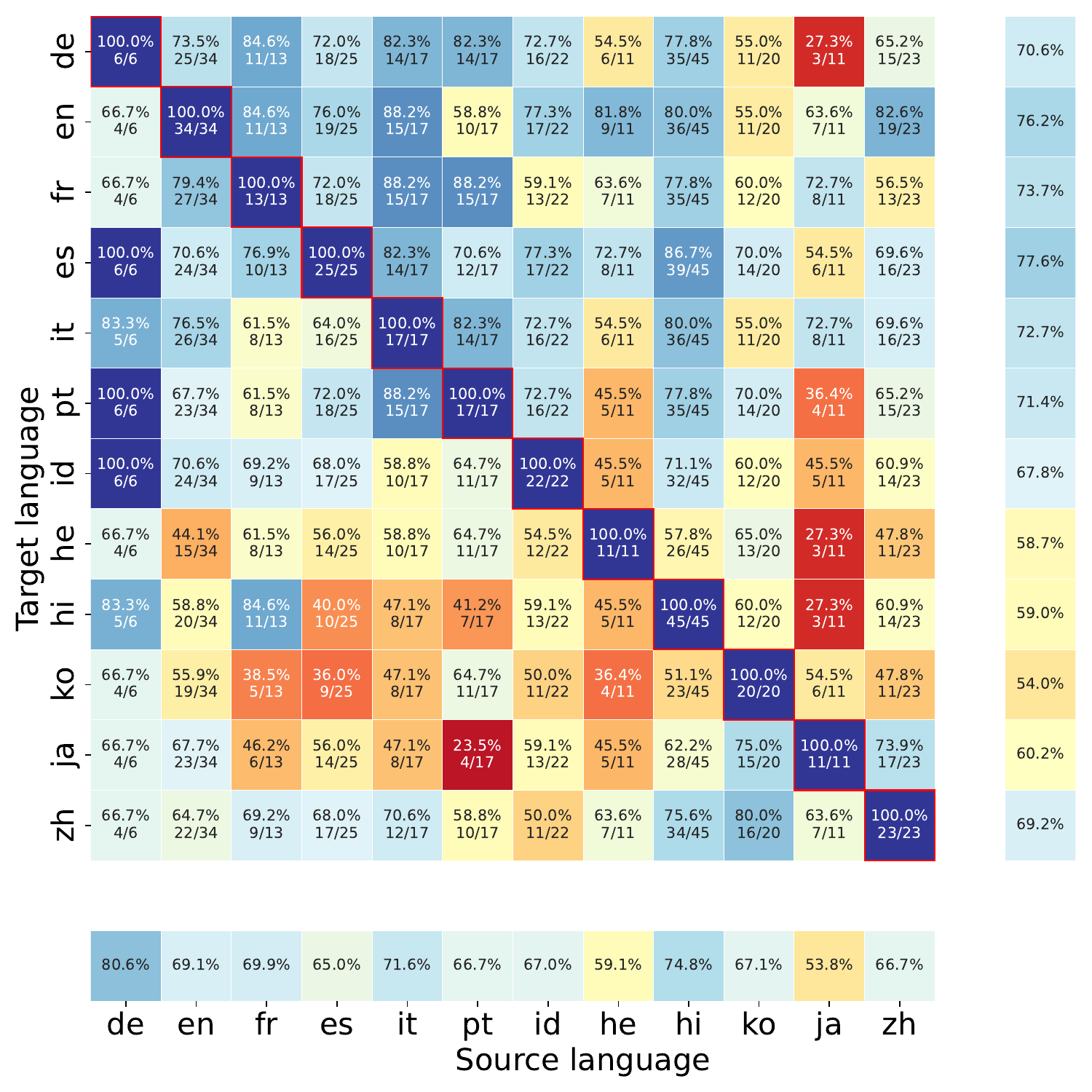}
  \end{center}
    \caption{\textit{Transfer score} results of Gemini 2.0 Pro broken down per source and target language. \textcolor{blue}{blue} is better, \textcolor{red}{red} is worse. Note the diagonal is perfect by the definition of the transfer metric.}
  \vspace{-30pt}
\label{fig:heatmap}
\end{wrapfigure}

\noindent Olmo,\footnote{\url{https://huggingface.co/allenai/OLMo-2-1124-7B-Instruct}} all in sizes of between 7 and 9 billion parameters. 
In addition, the black-box nature of \mbox{\eclektic} allows the evaluation of closed models. We therefore included also GPT 4o,\footnote{We used version \texttt{gpt-4o-2024-11-20}.} Claude 3.5 Sonnet,\footnote{Version \texttt{claude-3-5-sonnet-20241022}.} and Gemini 2.0 in Pro\footnote{Currently an experimental release.} and Flash\footnote{Version \texttt{gemini-2.0-flash-001}} versions.

All models were evaluated on the question on its own without an explicit instruction in the input. 

\begin{table}[t]
    \small
    \centering
    \begin{tabular}{c c c c c}
    \toprule
        \multirow{2}{*}{Model} & \multicolumn{2}{c}{LLM as a Judge} & \multicolumn{2}{c}{String Based} \\
        & Overall & Transfer & Overall & Transfer \\
        \midrule
        Gemini 2.0 Pro & \textbf{41.6$\pm$1.5} & 65.0$\pm$1.8 & \textbf{41.4$\pm$1.5} & \textbf{61.2$\pm$1.7} \\
        GPT 4o & 38.8$\pm$1.4 & \textbf{67.0$\pm$1.8} & 33.7$\pm$1.4 & 58.0$\pm$1.8 \\
        Gemini 2.0 Flash & 34.6$\pm$1.4 & 62.3$\pm$1.9 & 32.1$\pm$1.4 & 56.0$\pm$1.8 \\
        Claude 3.5 Sonnet & 34.4$\pm$1.4 & 60.8$\pm$1.9 & 24.9$\pm$1.3 & 49.1$\pm$1.9 \\
        \midrule
        Gemma 2 9B & \textbf{8.7$\pm$0.8} & \textbf{40.3$\pm$3.1} & \textbf{9.6$\pm$0.9} & \textbf{37.4$\pm$2.4} \\
        Mistral Nemo & 7.1$\pm$0.8 & \textbf{38.9$\pm$3.4} & \textbf{10.1$\pm$0.9} & \textbf{37.0$\pm$2.4} \\
        Qwen 2.5 7B & 2.8$\pm$0.5 & 23.5$\pm$3.7 & \textbf{9.7$\pm$0.9} & \textbf{36.2$\pm$2.4} \\
        Olmo 2 7B & 1.6$\pm$0.3 & 17.2$\pm$3.7 & 1.5$\pm$0.4 & 15.1$\pm$2.6 \\
        \bottomrule
    \end{tabular}
    \caption{Performance for all proprietary and open models over all examples in \mbox{\textsc{ECLeKTic}} in both metrics.}
    \label{tab:main_results}
\end{table}

\subsection{LLMs Struggle with Knowledge Transfer}
\label{sec:main_results}

The results can be found in \autoref{tab:main_results}. They show a clear gap in performance between the two groups, with proprietary models clearly outperforming the open ones by a very wide margin, probably due to their bigger size in terms of parameters (see also \autoref{sec:sizes_analysis}).

Gemini 2.0 Pro is the best performing model, as defined in terms of \textit{overall success}. It manages to answer correctly, in both source and target language, 41.6\% of the examples. This score reveals that there is still a clear room for improvement in knowledge retrieval and transfer in all models. 
In terms of \textit{transfer score}, the trends are pretty similar, with GPT 4o leading the pack. But even this more permissive metric shows that the model failed to transfer the knowledge of 37\% of the questions correctly answered in their source language.

When comparing the metrics based on an LLM as a judge to the ones based on string matching, it is evident that the trends are largely similar: closed models are better than open ones and the ranking is between the models is similar. It is also shows that the Gemini instance used as a judge did not particularly favor the judged Gemini models. In fact, the opposite may be true, as the non-Gemini models were favored by the LLM judge compared to their string based evaluation.

All in all, we conclude that \eclektic presents a significant challenge to modern LLMs despite their impressive abilities overall.

\subsection{Shared Script Eases Transfer}
\label{sec:per_lang_breakdown}

In order to provide further insights into the factors affecting knowledge-transfer, \autoref{fig:heatmap} details a per-language breakdown of our results in terms of \textit{transfer score} of our best performing model, Gemini 2.0 Pro. This analysis shows that the model's ability to transfer knowledge is highly dependent on the source and target language, with average scores ranging from 23.5\% (transfer from Portuguese to Japanese) to 100\% (from German to Indonesian).

Transfer is much higher between languages with the same script, as evident from high \textit{transfer score} between German, English, French, Spanish, Italian, Portuguese and Indonesian. Note that the latter is not genealogically related to the rest of the Latin-written languages, yet it performs on par with the others when serving as a source or a target to other Latin-written languages. The dependence on script can be also seen in the \textit{transfer scores} between Chinese and Japanese, especially when compared to the transfer of Japanese with other languages. This finding aligns with previous works in the literature on the importance of script to transfer \cite{malkin-etal-2022-balanced, mittal2023mokb6, ifergan2024beneath}.

Additionally, this analysis reveals an asymmetry in \textit{transfer scores} depending on the role of the language, as a source or a target. For example, knowledge seems to be easily transferable \textit{from} Hindi, mostly to Latin-written languages and Chinese, resulting in a macro-averaged \textit{transfer score} of 78.6\%. But the transfer \textit{to} Hindi is much worse -- only 59.6\% when Hindi is the target language.

\begin{figure}[t]
\begin{minipage}[t]{0.49\textwidth}
    \centering
    \adjustbox{valign=t}{
    \includegraphics[width=\linewidth]{sizes.pdf}
    }
    \captionof{figure}{Performance of Qwen 2.5 models in various sizes in terms of both \textit{overall success} and \textit{transfer score}. While transfer saturates at around 14B parameters, \textit{overall success} keeps improving with the increasing model size.}
    \label{fig:learning_curve}
\end{minipage}\hfill
\begin{minipage}[t]{0.49\textwidth}
    \centering
    \small
    \begin{tabular}[t]{l c c}
    \toprule
        Gemini 2.0 Flash & Overall & Transfer \\
        \midrule
        Closed-book & 34.5$\pm$1.4 & 62.3$\pm$1.9 \\
        General hint & 35.3$\pm$1.4 & 64.4$\pm$1.9 \\
        Source language name & 41.4$\pm$1.5 & 70.0$\pm$1.8 \\
        Source language title & 47.4$\pm$1.4 & 75.8$\pm$1.6 \\
        Cross-lingual open-book & 94.3$\pm$0.7 & 96.0$\pm$0.6 \\
        \bottomrule
    \end{tabular}
    \captionof{table}{Performance of Gemini 2.0 Flash when prompted with hints adding increasing amounts of information, from a general hint to use knowledge in another language to a cross-lingual open-book QA.}
    \label{tab:prompt_results}
\end{minipage}
\end{figure}

A similar breakdown for the \textit{overall success} metric can be found in \autoref{sec:overall-heatmap}.

\subsection{Bigger Isn't Necessarily Better}
\label{sec:sizes_analysis}

To further explore the role of model size in the ability to transfer knowledge, we evaluated all 7 model sizes of the Qwen 2.5 series, in sizes of 0.5, 1.5, 3, 7, 14, 32, and 73 billion parameters. We plotted the \textit{transfer} and \textit{overall scores} of these models in \autoref{fig:learning_curve}.

The difference in the curves of the transfer and the \textit{overall scores} is evident. The performance in terms of \textit{overall score} continues to improve with almost every increase in model size, and it seems plausible that an even bigger version of Qwen would have an even better performance. On the other hand, in terms of \textit{transfer score}, the performance improves rapidly with the increase in model size up until 14B parameters and then somewhat saturates, as more than a 5-fold increase in model size only give an improvement of about 7 percentage points in transfer. Taken together, this leads to the conclusion that the improvement in the \textit{overall scores} of the bigger model comes mostly from their ability to retrieve more facts in their source language while the proportion of facts transferred rises only marginally.

\subsection{Hinting LLMs Into Success Is Not Easy}
\label{sec:hints}

In the evaluation setting we examined until now -- where models are only given the question itself -- we found that models have a significant room for improvement. In order to characterize the pitfalls that make models fail and how to avoid them, we conducted an ablation study by incorporating more and more information into the prompt given to the model.
We experimented with Gemini 2.0 Flash in the following settings:
\begin{itemize}[nosep]
    \item \textit{Closed-book}. This is the setting used for the main experiment, giving the model only the question to be answered.
    \item \textit{General hint}. %
    In this setting an instruction is given to the model to answer the question while also instructing it to use its knowledge in other languages if it sees fit.
    \item \textit{Source language name}. This setting is very similar to the previous one but the name of the language with the relevant knowledge is given explicitly as part of the prompt.
    \item \textit{Source language title}. Here the model is given not only the name of the language but also the title of the Wikipedia article from which the question was generated. The title is given in its source language.
    \item \textit{Cross-lingual open-book}. In this last setting, the prompt includes the context in its source language and the question in the target language, so the model can completely disregard its parametric knowledge and it is only required to bridge the language differences in its input. This experiment is equivalent to that done by \citet{chua2024crosslingual}.
\end{itemize}
\noindent The prompts used in all of these settings are in \autoref{sec:prompts}.

The results, in \autoref{tab:prompt_results}, show that simply hinting the model towards cross-lingual knowledge is not enough and provides only insignificant improvement. Revealing the source language and the topic leads to a much improved performance of about 7 and 12 percentage points, respectively, indicating that in some cases the knowledge is partially available to the model, but requires some guidance. 

However, the most substantial improvement, almost to the point of solving \eclektic, comes when we gave the model 
the correct context, just in the source language. The model has no problem to reason across languages to produce the correct answer in 94.3\% of the examples. This means that when the knowledge is available in the prompt, transferring it is less of a problem. The limited performance over \eclektic is then more likely to arise from the difficulty in retrieving knowledge cross-lingually rather than processing it.

\section{Related Work}

\subsection{White-Boxing Methods for Transfer Evaluation}

This paper presents a QA dataset to evaluate cross-lingual knowledge transfer in a black-box manner without access to the models' inner workings. Another line of existing works aims to evaluate the same model behaviour by direct observation of the internal mechanisms of open models.

One line of works utilized knowledge editing to determine whether the model's representation of knowledge is causally linked across languages \cite{ifergan2024beneath, Wei2024MLaKEMK}. Other works used direct observation of activations in order to understand the extent to which transfer occurs \cite{chen2023journey, zhao2024large}. Beyond the usage of fairly complex mechanisms for transfer evaluation that may introduce confounding factors, the main drawback of these white-boxing methods is that they require access to the model's parameters and therefore can only be applied to open-source models and not to modern proprietary SOTA models.

\subsection{Types of Cross-Lingual Transfer}

The term \textit{cross-lingual transfer} has been extensively used in the NLP literature. The idea of one language benefiting from resources in another through a shared representation goes back to \citet{mcdonald-etal-2011-multi} at least. However, this term was used to refer to different specific experimental settings, some of them significantly different than the setting of this work.

The first distinction in the literature concerns \textit{what} is being transferred, specifically the difference between cross-lingual \textit{skill transfer} and \textit{knowledge transfer} Cross-lingual \textit{skill} transfer is the ability to generalize a given skill to unseen languages regardless of the language that was used to learn it. E.g. perfecting multilingual summarization when learning to summarize from English data, or excelling at multilingual instruction following while only exposed to a few languages  \citep{hu2020xtreme, turc2021revisiting, malkin-etal-2022-balanced, huang-etal-2023-languages, shaham-etal-2024-multilingual}. 
On the other hand, cross-lingual \textit{knowledge} transfer is the ability to retrieve factual knowledge from one language's data when queried with any language \citep{asai2020xor, limkonchotiwat2022cl, mittal2023mokb6, chua2024crosslingual, litschko2024cross}.

In addition, methods for evaluation of cross-lingual transfer can also be orthogonally 
categorized into two broad approaches: fine-tuning on sources, then testing on targets \cite[e.g.,][]{shaham-etal-2024-multilingual, limkonchotiwat2022cl} vs. evaluation only with the targets \cite[e.g.,][]{malkin-etal-2022-balanced, chua2024crosslingual}. While the earlier enables higher control over the data and experimental setting, the latter is less expensive and reflects how models behave in the wild.

Lastly, when dealing with cross-lingual knowledge transfer, the knowledge can be either parametric \cite[e.g.,][]{rajaee2024analyzing} or contextual \cite[e.g.,][]{chua2024crosslingual, mondshine2025english}, a distinction explored by \citet{neeman2022disentqa} in a monolingual setting. 

Within this taxonomy of cross-lingual transfer related works, \eclektic clearly belongs to methods evaluating parametric knowledge transfer without fine-tuning. In addition, the experiment in \autoref{sec:hints} gradually transition from parametric to contextual knowledge transfer, with the \textit{cross-lingual open-book} experiment occupying the other end of that spectrum.

\section{Conclusions}

In this paper we presented \eclektic, a closed-book QA evaluation set for measuring models' ability to transfer knowledge in their parametric memory across languages. This black-box evaluation is made possible by carefully phrasing questions that target topics highly visible in one language and not in any of the others. Our benchmark allows simple and reliable transfer evaluation which is, for the first time, easily applicable also to API-fenced models.
Our results show that cross-lingual knowledge transfer is a difficult task that is far from solved, and that \eclektic can be employed to indicate on the progress made towards consistent and inclusive multilingual language models.

\section{Limitations}

\paragraph{Time sensitivity} When constructing \eclektic we relied on the distribution of topics in Wikipedia in different languages as reflected in July 2023. Since then, and in the future, it is of course possible for the article distribution to change, mostly as new topics become more prominent for the speakers of a given language, on Wikipedia and in general. This makes \eclektic somewhat time-dependent, so in the future it would probably require an update.

\paragraph{Number of languages} Although the data \eclektic covers varied languages, its coverage is obviously partial. It covers only 6 out of the 10 most spoken languages according to Ethnologue,\footnote{\url{https://www.ethnologue.com/statistics/}} and 8 out of the 10 largest Wikipedias in terms of active users.\footnote{\url{https://en.wikipedia.org/wiki/List_of_Wikipedias}} However, we limited the number of languages to 12 for two reasons: (1) the annotation efforts took about 400 human hours already with 12 languages, and (2) we wanted to keep the benchmark light-weight and because of the exhaustive translation the final number of prompts for evaluation is quadratic in the number of languages. %

The limited number of languages also make the conclusions of \autoref{sec:per_lang_breakdown} less unequivocal, but the fact the previous works pointed to the same conclusions \cite{malkin-etal-2022-balanced, mittal2023mokb6, ifergan2024beneath} provides some reassurance.

\bibliography{custom,anthology_1,anthology_2}

\begin{thebibliography}{33}
\providecommand{\natexlab}[1]{#1}
\providecommand{\url}[1]{\texttt{#1}}
\expandafter\ifx\csname urlstyle\endcsname\relax
  \providecommand{\doi}[1]{doi: #1}\else
  \providecommand{\doi}{doi: \begingroup \urlstyle{rm}\Url}\fi

\bibitem[Asai et~al.(2020)Asai, Kasai, Clark, Lee, Choi, and Hajishirzi]{asai2020xor}
Akari Asai, Jungo Kasai, Jonathan~H Clark, Kenton Lee, Eunsol Choi, and Hannaneh Hajishirzi.
\newblock Xor qa: Cross-lingual open-retrieval question answering.
\newblock \emph{arXiv preprint arXiv:2010.11856}, 2020.

\bibitem[Brown et~al.(2020)Brown, Mann, Ryder, Subbiah, Kaplan, Dhariwal, Neelakantan, Shyam, Sastry, Askell, et~al.]{gpt3}
Tom Brown, Benjamin Mann, Nick Ryder, Melanie Subbiah, Jared~D Kaplan, Prafulla Dhariwal, Arvind Neelakantan, Pranav Shyam, Girish Sastry, Amanda Askell, et~al.
\newblock Language models are few-shot learners.
\newblock \emph{Advances in neural information processing systems}, 33:\penalty0 1877--1901, 2020.

\bibitem[Chen et~al.(2019)Chen, Stanovsky, Singh, and Gardner]{chen-etal-2019-evaluating}
Anthony Chen, Gabriel Stanovsky, Sameer Singh, and Matt Gardner.
\newblock Evaluating question answering evaluation.
\newblock In Adam Fisch, Alon Talmor, Robin Jia, Minjoon Seo, Eunsol Choi, and Danqi Chen (eds.), \emph{Proceedings of the 2nd Workshop on Machine Reading for Question Answering}, pp.\  119--124, Hong Kong, China, November 2019. Association for Computational Linguistics.
\newblock \doi{10.18653/v1/D19-5817}.
\newblock URL \url{https://aclanthology.org/D19-5817}.

\bibitem[Chen et~al.(2024)Chen, Cao, Chen, Liu, and Zhao]{chen2023journey}
Yuheng Chen, Pengfei Cao, Yubo Chen, Kang Liu, and Jun Zhao.
\newblock Journey to the center of the knowledge neurons: Discoveries of language-independent knowledge neurons and degenerate knowledge neurons.
\newblock In \emph{Proceedings of the AAAI Conference on Artificial Intelligence}, volume~38, pp.\  17817--17825, 2024.

\bibitem[Chiang \& Lee(2023)Chiang and Lee]{chiang-lee-2023-large}
Cheng-Han Chiang and Hung-yi Lee.
\newblock Can large language models be an alternative to human evaluations?
\newblock In Anna Rogers, Jordan Boyd-Graber, and Naoaki Okazaki (eds.), \emph{Proceedings of the 61st Annual Meeting of the Association for Computational Linguistics (Volume 1: Long Papers)}, pp.\  15607--15631, Toronto, Canada, July 2023. Association for Computational Linguistics.
\newblock \doi{10.18653/v1/2023.acl-long.870}.
\newblock URL \url{https://aclanthology.org/2023.acl-long.870}.

\bibitem[Chowdhery et~al.(2023)Chowdhery, Narang, Devlin, Bosma, Mishra, Roberts, Barham, Chung, Sutton, Gehrmann, et~al.]{palm}
Aakanksha Chowdhery, Sharan Narang, Jacob Devlin, Maarten Bosma, Gaurav Mishra, Adam Roberts, Paul Barham, Hyung~Won Chung, Charles Sutton, Sebastian Gehrmann, et~al.
\newblock Palm: Scaling language modeling with pathways.
\newblock \emph{Journal of Machine Learning Research}, 24\penalty0 (240):\penalty0 1--113, 2023.

\bibitem[Chua et~al.(2024)Chua, Ghazi, Huang, Kamath, Kumar, Manurangsi, Sinha, Xie, and Zhang]{chua2024crosslingual}
Lynn Chua, Badih Ghazi, Yangsibo Huang, Pritish Kamath, Ravi Kumar, Pasin Manurangsi, Amer Sinha, Chulin Xie, and Chiyuan Zhang.
\newblock Crosslingual capabilities and knowledge barriers in multilingual large language models.
\newblock In \emph{NeurIPS 2024 Workshop on Compositional Learning: Perspectives, Methods, and Paths Forward}, 2024.

\bibitem[{Gemini Team}(2024)]{gemini}
{Gemini Team}.
\newblock Gemini: A family of highly capable multimodal models, 2024.
\newblock URL \url{https://arxiv.org/abs/2312.11805}.

\bibitem[Hu et~al.(2020)Hu, Ruder, Siddhant, Neubig, Firat, and Johnson]{hu2020xtreme}
Junjie Hu, Sebastian Ruder, Aditya Siddhant, Graham Neubig, Orhan Firat, and Melvin Johnson.
\newblock Xtreme: A massively multilingual multi-task benchmark for evaluating cross-lingual generalisation.
\newblock In \emph{International Conference on Machine Learning}, pp.\  4411--4421. PMLR, 2020.

\bibitem[Huang et~al.(2023)Huang, Tang, Zhang, Zhao, Song, Xia, and Wei]{huang-etal-2023-languages}
Haoyang Huang, Tianyi Tang, Dongdong Zhang, Xin Zhao, Ting Song, Yan Xia, and Furu Wei.
\newblock Not all languages are created equal in {LLM}s: Improving multilingual capability by cross-lingual-thought prompting.
\newblock In Houda Bouamor, Juan Pino, and Kalika Bali (eds.), \emph{Findings of the Association for Computational Linguistics: EMNLP 2023}, pp.\  12365--12394, Singapore, December 2023. Association for Computational Linguistics.
\newblock \doi{10.18653/v1/2023.findings-emnlp.826}.
\newblock URL \url{https://aclanthology.org/2023.findings-emnlp.826}.

\bibitem[Ifergan et~al.(2024)Ifergan, Choshen, Aharoni, Szpektor, and Abend]{ifergan2024beneath}
Maxim Ifergan, Leshem Choshen, Roee Aharoni, Idan Szpektor, and Omri Abend.
\newblock Beneath the surface of consistency: Exploring cross-lingual knowledge representation sharing in llms, 2024.
\newblock URL \url{https://arxiv.org/abs/2408.10646}.

\bibitem[Jiang et~al.(2020)Jiang, Anastasopoulos, Araki, Ding, and Neubig]{jiang-etal-2020-x}
Zhengbao Jiang, Antonios Anastasopoulos, Jun Araki, Haibo Ding, and Graham Neubig.
\newblock {X}-{FACTR}: Multilingual factual knowledge retrieval from pretrained language models.
\newblock In Bonnie Webber, Trevor Cohn, Yulan He, and Yang Liu (eds.), \emph{Proceedings of the 2020 Conference on Empirical Methods in Natural Language Processing (EMNLP)}, pp.\  5943--5959, Online, November 2020. Association for Computational Linguistics.
\newblock \doi{10.18653/v1/2020.emnlp-main.479}.
\newblock URL \url{https://aclanthology.org/2020.emnlp-main.479}.

\bibitem[Kassner et~al.(2021)Kassner, Dufter, and Sch{\"u}tze]{kassner-etal-2021-multilingual}
Nora Kassner, Philipp Dufter, and Hinrich Sch{\"u}tze.
\newblock Multilingual {LAMA}: Investigating knowledge in multilingual pretrained language models.
\newblock In Paola Merlo, Jorg Tiedemann, and Reut Tsarfaty (eds.), \emph{Proceedings of the 16th Conference of the European Chapter of the Association for Computational Linguistics: Main Volume}, pp.\  3250--3258, Online, April 2021. Association for Computational Linguistics.
\newblock \doi{10.18653/v1/2021.eacl-main.284}.
\newblock URL \url{https://aclanthology.org/2021.eacl-main.284}.

\bibitem[Limkonchotiwat et~al.(2022)Limkonchotiwat, Ponwitayarat, Udomcharoenchaikit, Chuangsuwanich, and Nutanong]{limkonchotiwat2022cl}
Peerat Limkonchotiwat, Wuttikorn Ponwitayarat, Can Udomcharoenchaikit, Ekapol Chuangsuwanich, and Sarana Nutanong.
\newblock Cl-relkt: Cross-lingual language knowledge transfer for multilingual retrieval question answering.
\newblock In \emph{Findings of the Association for Computational Linguistics: NAACL 2022}, pp.\  2141--2155, 2022.

\bibitem[Litschko et~al.(2024)Litschko, Kraus, Blaschke, and Plank]{litschko2024cross}
Robert Litschko, Oliver Kraus, Verena Blaschke, and Barbara Plank.
\newblock Cross-dialect information retrieval: Information access in low-resource and high-variance languages.
\newblock \emph{arXiv preprint arXiv:2412.12806}, 2024.

\bibitem[{Llama Team}(2024)]{llama}
{Llama Team}.
\newblock The llama 3 herd of models, 2024.
\newblock URL \url{https://arxiv.org/abs/2407.21783}.

\bibitem[Malkin et~al.(2022)Malkin, Limisiewicz, and Stanovsky]{malkin-etal-2022-balanced}
Dan Malkin, Tomasz Limisiewicz, and Gabriel Stanovsky.
\newblock A balanced data approach for evaluating cross-lingual transfer: Mapping the linguistic blood bank.
\newblock In Marine Carpuat, Marie-Catherine de~Marneffe, and Ivan~Vladimir Meza~Ruiz (eds.), \emph{Proceedings of the 2022 Conference of the North American Chapter of the Association for Computational Linguistics: Human Language Technologies}, pp.\  4903--4915, Seattle, United States, July 2022. Association for Computational Linguistics.
\newblock \doi{10.18653/v1/2022.naacl-main.361}.
\newblock URL \url{https://aclanthology.org/2022.naacl-main.361}.

\bibitem[McDonald et~al.(2011)McDonald, Petrov, and Hall]{mcdonald-etal-2011-multi}
Ryan McDonald, Slav Petrov, and Keith Hall.
\newblock Multi-source transfer of delexicalized dependency parsers.
\newblock In Regina Barzilay and Mark Johnson (eds.), \emph{Proceedings of the 2011 Conference on Empirical Methods in Natural Language Processing}, pp.\  62--72, Edinburgh, Scotland, UK., July 2011. Association for Computational Linguistics.
\newblock URL \url{https://aclanthology.org/D11-1006}.

\bibitem[Mittal et~al.(2023)Mittal, Kolluru, Chakrabarti, et~al.]{mittal2023mokb6}
Shubham Mittal, Keshav Kolluru, Soumen Chakrabarti, et~al.
\newblock mokb6: A multilingual open knowledge base completion benchmark.
\newblock In \emph{Proceedings of the 61st Annual Meeting of the Association for Computational Linguistics (Volume 2: Short Papers)}, pp.\  201--214, 2023.

\bibitem[Mondshine et~al.(2025)Mondshine, Paz-Argaman, and Tsarfaty]{mondshine2025english}
Itai Mondshine, Tzuf Paz-Argaman, and Reut Tsarfaty.
\newblock Beyond english: The impact of prompt translation strategies across languages and tasks in multilingual llms.
\newblock In \emph{Findings of the Association for Computational Linguistics: NAACL 2025}. Association for Computational Linguistics, June 2025.
\newblock URL \url{https://arxiv.org/abs/2502.09331}.

\bibitem[Neeman et~al.(2022)Neeman, Aharoni, Honovich, Choshen, Szpektor, and Abend]{neeman2022disentqa}
Ella Neeman, Roee Aharoni, Or~Honovich, Leshem Choshen, Idan Szpektor, and Omri Abend.
\newblock Disentqa: Disentangling parametric and contextual knowledge with counterfactual question answering.
\newblock \emph{arXiv preprint arXiv:2211.05655}, 2022.

\bibitem[Ohmer et~al.(2023)Ohmer, Bruni, and Hupkes]{ohmer-etal-2023-separating}
Xenia Ohmer, Elia Bruni, and Dieuwke Hupkes.
\newblock Separating form and meaning: Using self-consistency to quantify task understanding across multiple senses.
\newblock In Sebastian Gehrmann, Alex Wang, Jo{\~a}o Sedoc, Elizabeth Clark, Kaustubh Dhole, Khyathi~Raghavi Chandu, Enrico Santus, and Hooman Sedghamiz (eds.), \emph{Proceedings of the Third Workshop on Natural Language Generation, Evaluation, and Metrics (GEM)}, pp.\  258--276, Singapore, December 2023. Association for Computational Linguistics.
\newblock URL \url{https://aclanthology.org/2023.gem-1.22}.

\bibitem[OpenAI(2024)]{openai2024gpt4}
OpenAI.
\newblock Gpt-4 technical report, 2024.
\newblock URL \url{https://arxiv.org/abs/2303.08774}.

\bibitem[Qi et~al.(2023)Qi, Fern{\'a}ndez, and Bisazza]{qi-etal-2023-cross}
Jirui Qi, Raquel Fern{\'a}ndez, and Arianna Bisazza.
\newblock Cross-lingual consistency of factual knowledge in multilingual language models.
\newblock In Houda Bouamor, Juan Pino, and Kalika Bali (eds.), \emph{Proceedings of the 2023 Conference on Empirical Methods in Natural Language Processing}, pp.\  10650--10666, Singapore, December 2023. Association for Computational Linguistics.
\newblock \doi{10.18653/v1/2023.emnlp-main.658}.
\newblock URL \url{https://aclanthology.org/2023.emnlp-main.658}.

\bibitem[{Qwen Team}(2025)]{qwen25}
{Qwen Team}.
\newblock Qwen2.5 technical report, 2025.
\newblock URL \url{https://arxiv.org/abs/2412.15115}.

\bibitem[Rajaee \& Monz(2024)Rajaee and Monz]{rajaee2024analyzing}
Sara Rajaee and Christof Monz.
\newblock Analyzing the evaluation of cross-lingual knowledge transfer in multilingual language models.
\newblock \emph{arXiv preprint arXiv:2402.02099}, 2024.

\bibitem[Shaham et~al.(2024)Shaham, Herzig, Aharoni, Szpektor, Tsarfaty, and Eyal]{shaham-etal-2024-multilingual}
Uri Shaham, Jonathan Herzig, Roee Aharoni, Idan Szpektor, Reut Tsarfaty, and Matan Eyal.
\newblock Multilingual instruction tuning with just a pinch of multilinguality.
\newblock In Lun-Wei Ku, Andre Martins, and Vivek Srikumar (eds.), \emph{Findings of the Association for Computational Linguistics: ACL 2024}, pp.\  2304--2317, Bangkok, Thailand, August 2024. Association for Computational Linguistics.
\newblock \doi{10.18653/v1/2024.findings-acl.136}.
\newblock URL \url{https://aclanthology.org/2024.findings-acl.136}.

\bibitem[Turc et~al.(2021)Turc, Lee, Eisenstein, Chang, and Toutanova]{turc2021revisiting}
Iulia Turc, Kenton Lee, Jacob Eisenstein, Ming-Wei Chang, and Kristina Toutanova.
\newblock Revisiting the primacy of english in zero-shot cross-lingual transfer.
\newblock \emph{arXiv preprint arXiv:2106.16171}, 2021.

\bibitem[Verma et~al.(2024)Verma, Rassin, Das, Bhatt, Seshadri, Shah, Bilmes, Hajishirzi, and Elazar]{verma2024van}
Sahil Verma, Royi Rassin, Arnav Das, Gantavya Bhatt, Preethi Seshadri, Chirag Shah, Jeff Bilmes, Hannaneh Hajishirzi, and Yanai Elazar.
\newblock How many van goghs does it take to van gogh? finding the imitation threshold, 2024.
\newblock URL \url{https://arxiv.org/abs/2410.15002}.

\bibitem[Wei et~al.(2024)Wei, Deng, Pang, Ding, Shen, and Cheng]{Wei2024MLaKEMK}
Zihao Wei, Jingcheng Deng, Liang Pang, Hanxing Ding, Huawei Shen, and Xueqi Cheng.
\newblock Mlake: Multilingual knowledge editing benchmark for large language models.
\newblock \emph{ArXiv}, abs/2404.04990, 2024.
\newblock URL \url{https://api.semanticscholar.org/CorpusID:269005017}.

\bibitem[Zhao et~al.(2024)Zhao, Zhang, Chen, Kawaguchi, and Bing]{zhao2024large}
Yiran Zhao, Wenxuan Zhang, Guizhen Chen, Kenji Kawaguchi, and Lidong Bing.
\newblock How do large language models handle multilingualism?, 2024.
\newblock URL \url{https://arxiv.org/abs/2402.18815}.

\bibitem[Zheng et~al.(2023)Zheng, Chiang, Sheng, Zhuang, Wu, Zhuang, Lin, Li, Li, Xing, et~al.]{zheng2023judging}
Lianmin Zheng, Wei-Lin Chiang, Ying Sheng, Siyuan Zhuang, Zhanghao Wu, Yonghao Zhuang, Zi~Lin, Zhuohan Li, Dacheng Li, Eric Xing, et~al.
\newblock Judging llm-as-a-judge with mt-bench and chatbot arena.
\newblock \emph{Advances in Neural Information Processing Systems}, 36:\penalty0 46595--46623, 2023.

\bibitem[Zhou et~al.(2025)Zhou, Arnold, Ding, Weinberger, Hua, and Sha]{zhou2025graders}
Jin~Peng Zhou, Sébastien M.~R. Arnold, Nan Ding, Kilian~Q. Weinberger, Nan Hua, and Fei Sha.
\newblock Graders should cheat: privileged information enables expert-level automated evaluations, 2025.
\newblock URL \url{https://arxiv.org/abs/2502.10961}.

\end{thebibliography}
\bibliographystyle{iclr2026_conference}

\appendix

\section{Annotation Guidelines}
\label{sec:guidelines}

During the creation of \eclektic human annotators worked in two stages based on outputs of Gemini (see \autoref{sec:prompts}). Here we provide the annotation guidelines there were given in each stage.

\subsection{In-Language Verification and Filtering}

\textbf{Purpose}

We would like to examine language models’ (LMs) ability to answer questions in a language while seeing little relevant information in that language. For that we scraped a set of Wikipedia pages that appear only in one language and not in the others. The desired set should contain questions like:

Q: “Em qual ano foi criada a Fundação Zoobotânica de Belo Horizonte?” (“In what year was the Zoobotanical Foundation of Belo Horizonte created?”)

We assume that the model has seen the answer to this question in pt-BR, but might not have seen it in other languages, for example ja-JP. We’d like to test the model’s ability to answer this question correctly, when prompted in all other 12 languages. For that it is critical to have a solid set of questions, answers, contexts that are suitable for this purpose.

We would like to ask for your help with creating that dataset.

\textbf{The Process}

The data includes contexts from Wikipedia articles (specifically, the 10 first sentences of each article), and a pair of suggested question and answer based on each context. You need to filter irrelevant examples and modify or rewrite questions and answers that are faulty due to the automatic generation or other reasons.

In general, make sure that the question and its answer are clear, written in proper language. Check whether the question is answerable based on the given article and is faithful to the information conveyed in the article, but also keep in mind that the question should be understandable and answerable even without the given context (for example, by searching the web).

Whenever unsure of how to solve problems with a specific example, feel free to discard the example (i.e., select "to remove" in column D). We want to end up with a high quality dataset, even if slightly smaller.

Specifically, we need you to attend the following potential issues:
\begin{itemize}[nosep]
\item Make sure that the Wikipedia article that the example is based on is a contentful article, that is not a disambiguation page or a page dedicated to a list of any kind. If it is, mark the example as “to remove” in column D.
\item Make sure that the topic of the article and/or the question relates to the speakers of that relevant language. In other words, mark as “to remove” also articles that relate to general topics like science or sociology. Note that a question in, e.g., German about a German biologist counts as a question that relates to the language and not a general scientific question, but a question about natural phenomena – should be removed.\\
Also note that questions in one language, that refer to topics relevant to another language, are also to be removed. For example, a question in Chinese that refer to a K-pop band that is famous mostly in Korea.
\item Make sure that the question is answerable without reference to the content of the article and that it does not include an explicit refernce to the context (e.g., "based on the given paragraph, who was...?"). If this issue is easily fixable, suggest a better phrasing of the question in column E, else -- mark it to remove in column D.
\item Remember that he question are to be translated to other languages so they should be understandable universally. That is to say, that questions that refer to locally-known figures add the relevant information explicitly to the question. For example, when a question is asking about a TV series in India, prepend "the Indian TV series" to its name. You can use Google search to understand whether an entity is locally- or globally-known.\\
The same kind of disambiguation is needed when referring to certain institutions and other named entities. For example, in a Hebrew question, disambiguate "the supreme court" to "the Israeli supreme court".
\end{itemize}

\subsection{Translation Verification and Modification}

\textbf{Purpose}

We would like to examine language models’ (LMs) ability to answer questions in a language while seeing little relevant information in that language. For that we constructed a set of locale-relevant questions, assuming that the model saw information mostly in the language relevant to the locale and not in other languages.

Example:
Q: “Em qual ano foi criada a Fundação Zoobotânica de Belo Horizonte?” (“In what year was the Zoobotanical Foundation of Belo Horizonte created?”)

We assume that the model has seen the answer to this question in pt-BR, but might not have seen it in other languages, for example ja-JP. We’d like to test the model’s ability to answer this question correctly, when prompted in all other 12 languages. For that purpose, we automatically translated them to other (target) languages.

We would like to ask for your help with verifying the translations of the questions and with making sure the questions are indeed on subjects related to the language they are presented in.

\textbf{The Process}

The data includes factual questions in 12 languages: German, English, Spanish, French, Hebrew, Hindi, Indonesian, Italian, Japanese, Korean, Portuguese, Chinese (Mandarin), and a translation of each question and answer pair into all other 11 languages, which we’d like to verify, and sometimes slightly fix, if it’s not very natural or doesn’t make sense (see below). 

In order to bypass translation between uncommon language pairs, the translation verification will be done in 2 stages:

\begin{itemize}[nosep]
    \item Verify and correct translations from each of the languages to English, which can be done by a native speaker of each language;
    \item Verify the translation to all other languages, using the now-gold English translation as a reference. Doing so will require every verifier to be knowledgeable only in one language and in English.
\end{itemize}

\textbf{Phase 1 Task}

In this 1st phase, you only need to verify translations from a given language to English. In cases where the question is generally fine but has some minor issues, you could modify the English translation and, if needed, also the original question. If the fix is impossible or includes a complete rewrite of the question or answer, you can also recommend dropping that question altogether.

\textit{Possible Corrections Needed}

The automatic translation may have caused multiple problems. Common issues might include unnatural output and missing or added information. In addition, there may be several problems specific to the nature of questions we have at hand, like transliteration of uncommon named entities.

\begin{enumerate}[nosep]
    \item Missing information: In some cases the question is perfectly understandable in the original language, but the translation to English could use a clarification. In these cases modify the original question as well as its translation. \\
    For example, in Spanish, the question: ¿Cuál es la primera de las Siete Palabras? (“what is the first of the Seven Words?”) May be perfectly understandable but in other languages, specifically those with no Christian tradition, it isn’t clear what words are referred to. In this case modify the Spanish original and English translation to clarify that these are the last seven words of Jesus on the cross.\\
    Another example, in Hebrew, the question: \cjRL{my mglm 't dmwtw /sl db--b`r h.s`yr bsdrh "ml'K m/s.hyt"?} (“Who plays the character of young Dov-Ber in the series ‘Angel of Destruction’?”) is missing an explicit description of the series as an Israeli series, which may make the question somewhat less understandable in other languages.
    \item Improper transliteration: Sometimes, transliteration depends on some external factors that are missed by the automatic translator. In these cases, the English translation should be amended.\\
    \begin{CJK}{UTF8}{gbsn}
    For example, the Chinese question 李显龙的生日是哪一天？refers to 李显龙 who is a Singaporean politician. If the English translation is “when was Li Xianlong born?” it should be amended to reflect the fact that proper transliteration of Chinese names from Singapore does not use the Pinyin system. So the English translation should be modified to “when was Lee Hsien Loong born?”
    \end{CJK}
    \item Questions that are not locale-specific: If you encounter a question that is not specifically relevant to the source language, please select “Recommend removing” in column D. Such questions can be too general, not specific to any language, for example “How many cm are in an inch?” a question about a Korean actor when the source language is French. If it is unclear if a question is specifically relevant to the source language or not, please do not recommend removing it.
    \item Translatable titles: Names of movies, TV series, books and the like, are a major possible pitfall since they tend to be a translatable phrase and the treatment of uncommon titles is not trivial if they don’t have an official translation. In these cases, we would like to make sure that such official translation indeed does not exist and if so, translate the name as best as possible and make sure it is clearly marked with “...” or «…» or any other way common in the target language.
\end{enumerate}

\textbf{Phase 2}

In the 2nd phase we would like you to verify the translation of the questions from English to your language of expertise. Most questions were translated to English from another language (see phase 1), in these cases you could also take into account the original question, in case you know the source language. 
In this phase you could only amend or verify the non-English translation. Do not amend the English nor the source language. For common issues that may require amendment of the translation, see phase 1. In cases of severe amendments needed in the source language that were missed in previous iterations, you could also flag the question for removal by selecting “Recommend removing” in column D.

\section{Prompts}
\label{sec:prompts}

\subsection{\eclektic Creation}

During the creation of \eclektic, LLMs were used for question and answer generation and for the translation of the examples from their respective source languages to all other target languages. Human annotators later verified and fixed their outputs (see \autoref{sec:guidelines}). We provide prompts for QA generation, translation, LLM as a judge and our ablation prompts in \autoref{table:QA_pairs_prompt}, \autoref{table:translation_prompt}, \autoref{table:judging_prompt} \autoref{table:ablation_prompt} respectively.

\begin{figure*}[htb]
  \centering
  \fbox{\begin{minipage}{\textwidth}
  \small
    \textbf{Task:} Formulate a question in TARGET\_LANGUAGE that requires a deep understanding of a given SOURCE\_LANGUAGE Wikipedia paragraph.
    
    \bigskip
    
    \textbf{Requirements:}
    
    \textasteriskcentered{} \textbf{Context-Specific:} The question must be answerable solely through information presented within the paragraph, excluding general knowledge or common sense.
    
    \textasteriskcentered{} \textbf{Self-Contained:}  The question should be completely self-explanatory, providing all necessary context within its phrasing.  Assume the reader has no access to the paragraph when answering the question.

    \textasteriskcentered{} \textbf{Single Concrete Factual Detail:}

- The question should not require multiple answers or involve listing multiple details.

- Avoid asking about opinions, interpretations.

- In you can't answer the question, prefer to generate another question.

- Focus on extracting a specific, concrete, factual detail that the paragraph directly states.

- Be specific:

  \hspace{1em} - If you are asking about an entity be clear about it -- Use full names for example.
  
  \hspace{1em} - Mention expected granularity: If you are asking about a date, instead of asking "when", ask for a decade, year, month, date etc. If you are asking about a location, instead of asking "where", ask for a country, state, city, street, landmark etc.
  
- Avoid asking questions that their answers are acronyms.

  Even for non-English examples keep the convention of using the English words "Paragraph", "Response", "question" and "answer" for specifying the parts being generated.
  
  Generate only the question and answer. No need to continue with additional examples.
  
\bigskip
  
  \textbf{Examples:}

  Paragraph:
  The Great Barrier Reef is the world's largest coral reef system, composed of over 2,900 individual reefs and 900 islands stretching for over 2,300 kilometers (1,400 mi) over an area of approximately 344,400 square kilometers (133,000 sq mi). The reef is located in the Coral Sea, off the coast of Queensland, Australia. The Great Barrier Reef can be seen from outer space and is the world's biggest single structure made by living organisms.

  Response:
  
  Question: Where is the Great Barrier Reef located?
  Answer: Coral Sea, off the coast of Queensland, Australia
  
  \bigskip
  
  Paragraph:
  Die Cazoo Snookerweltmeisterschaft 2023 wurde vom 15. April bis 1. Mai im Crucible Theatre in Sheffield ausgetragen. Mit ihr endete die Saison 2022/23 der World Snooker Tour.[1] Titelverteidiger Ronnie O'Sullivan scheiterte im Viertelfinale gegen Luca Brecel. Der Belgier erreichte das Finale und schlug dort den vierfachen Weltmeister Mark Selby mit 18:15. Brecel ist damit der erste Kontinentaleuropäer, der Weltmeister wurde. In diesem Jahr wurden noch weitere Bestmarken in Bezug auf die 47-jährige „Crucible-Ära“ aufgestellt. Unter anderem übertraf Ronnie O'Sullivan mit seiner 31. Endrundenteilnahme die 30 Teilnahmen von Steve Davis.[2] O'Sullivan erzielte auch sein 200. WM-Century-Break. Zweimal wurde ein Maximum Break erzielt, was es 2008 bereits einmal gegeben hatte; das „perfekte Break“ in einem WM-Finale gelang 2023 erstmals Mark Selby.

  Response:
  
  Question: Gegen wen verlor Ronnie O'Sullivan im Viertelfinale der Snooker-Weltmeisterschaft 2023?
  Answer: Luca Brecel
  
  \bigskip
  
    Paragraph:
  TEXT

  Response:
  
  \end{minipage}}
  \caption{The prompt that was used to generate question and answers given a context. In the prompt we first start by explaining the task, followed by 2 examples and then allowing placeholders for passing the text for which we'd like to generate QA pairs. Capitalized text should be replaced with required source and target languages and text input.}
  \label{table:QA_pairs_prompt}
\end{figure*}

\begin{figure*}[htb]
  \centering
  \fbox{\begin{minipage}{\textwidth}

  Translate the provided text from IN\_LANG to OUT\_LANG while maintaining the original meaning and intent. Ensure accuracy and preserve the entities and concepts expressed in the source text.
\\~\\
  Input in IN\_LANG:

  TEXT
 \\~\\
  Response in OUT\_LANG:

  \end{minipage}}
  \caption{The prompt that was used for the translation step. We start by explaining the task followed by the text we'd like to translate. Capitalized text should be replaced with required input and output languages and text input.}
  \label{table:translation_prompt}
\end{figure*}

\begin{figure*}[htb]
  \centering
  \fbox{\begin{minipage}{\textwidth}

  \textbf{Task:} Determine if an answer to the question is supported by a given text.
\\~\\
  \textbf{Input (in TARGET\_LANG):}
  
  - Text
  
  - Question
  
  - Answer
\\~\\
  \textbf{Single Word Output (in English):}
  
  - YES: Answer is derived from the text.
  
  - NO: Answer is not derived from the text.
\\~\\
  Text:
  
  TEXT
\\~\\
  Question:
  
  QUESTION
\\~\\
  Answer:
  
  PREDICTED\_ANSWER
\\~\\
  Output:

  \end{minipage}}
  \caption{Our judging prompt was used to classify whether a predicted answer is inferred given the Wikipedia text that was used to generate the QA pairs. Once again we replace during running time the capitalized text with the example language, text, question and predicted answer}
  \label{table:judging_prompt}
\end{figure*}

\begin{figure*}[htb]
  \centering
  \fbox{\begin{minipage}{\textwidth}

\textbf{Closed-book}

QUESTION
\\~\\
\textbf{General hint}

Answer the following question based on your knowledge in another language.

QUESTION
\\~\\
\textbf{Source language name}

Answer the following question based on your knowledge in \{in\_lang\}.

QUESTION
\\~\\
\textbf{Source language title}

Answer the following question based on your knowledge in \{in\_lang\} about \{original\_title\}.

QUESTION
\\~\\
\textbf{Cross-lingual open-book}

Context: SOURCE\_LANG\_CONTEXT

QUESTION

  \end{minipage}}
  \caption{Prompts for ablation experiments in \autoref{sec:hints}}
  \label{table:ablation_prompt}
\end{figure*}

\noindent\textbf{} 

\begin{figure*}[t]
    \centering
    \includegraphics[width=\columnwidth]{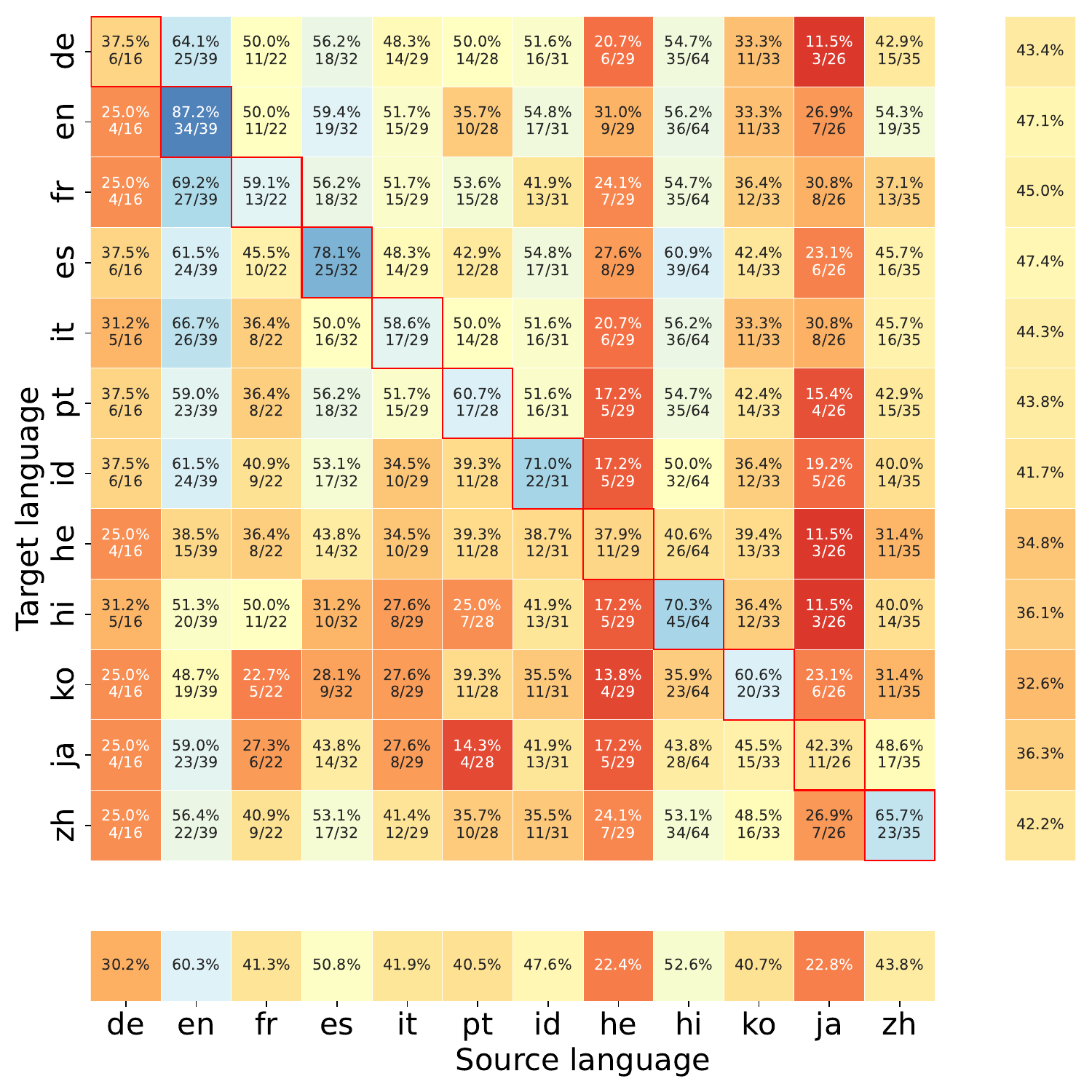} %
    \caption{\textit{Overall success} results of Gemini 2.0 Pro broken down per source and target language. \textcolor{blue}{blue} is better, \textcolor{red}{red} is worse. Note the diagonal is perfect by the definition of the transfer metric.}
    \label{fig:heatmap-overall}
\end{figure*}

\section{Per-Language Breakdown - Overall Success}
\label{sec:overall-heatmap}

\autoref{fig:heatmap-overall} contains per-language breakdown of the results in terms of \textit{overall success} of the best performing model, Gemini 2.0 Pro, similar to the \textit{transfer score} breakdown done in \autoref{sec:per_lang_breakdown}. It shows that when taking into account the ability to answer questions in their source language, some languages lead to worse performance when used as a source. For example, the low scores on the diagonal for German, Hebrew and Japanese lead to worse scores on the entire respective columns.

\section{Per-Language Domain Distribution}
\label{sec:per-language-domains}

\begin{figure*}[htbp]
    \centering
    \begin{subfigure}[b]{0.48\columnwidth}
        \includegraphics[width=\textwidth]{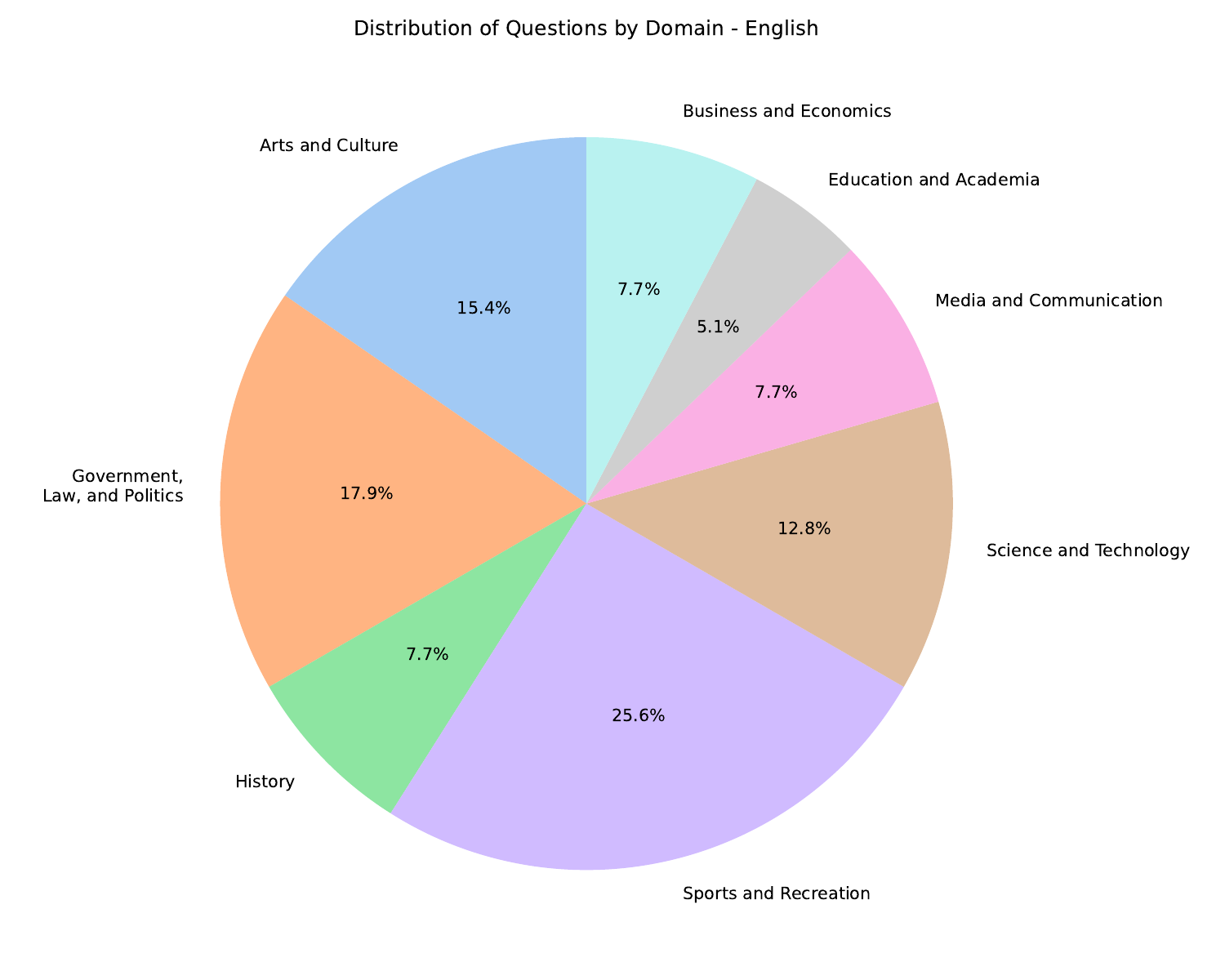}
        \label{fig:sub1}
    \end{subfigure}%
    \hfill
    \begin{subfigure}[b]{0.48\columnwidth}
        \includegraphics[width=\textwidth]{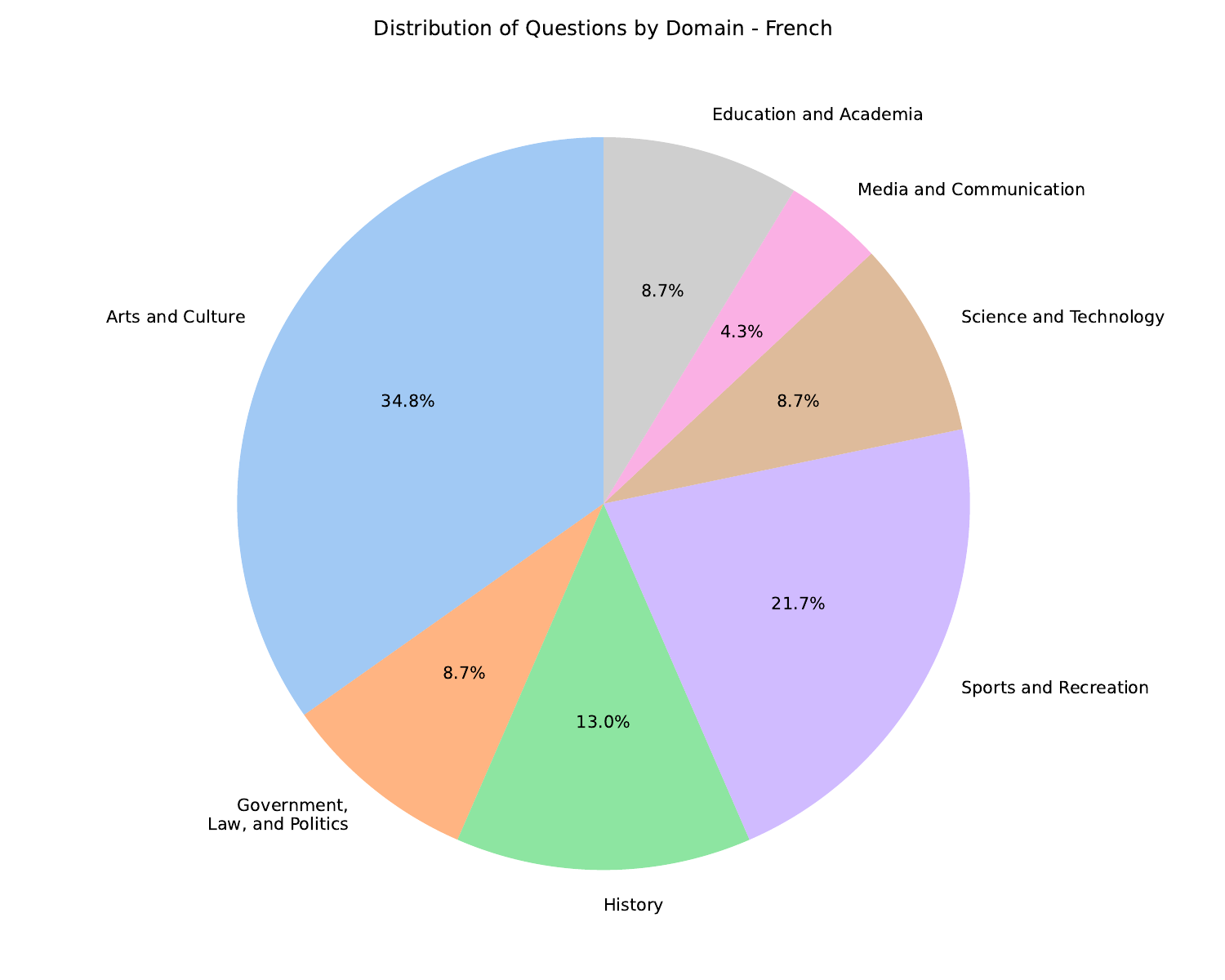}
        \label{fig:sub2}
    \end{subfigure}
    
    \vspace{\baselineskip}

    \begin{subfigure}[b]{0.48\columnwidth}
        \includegraphics[width=\textwidth]{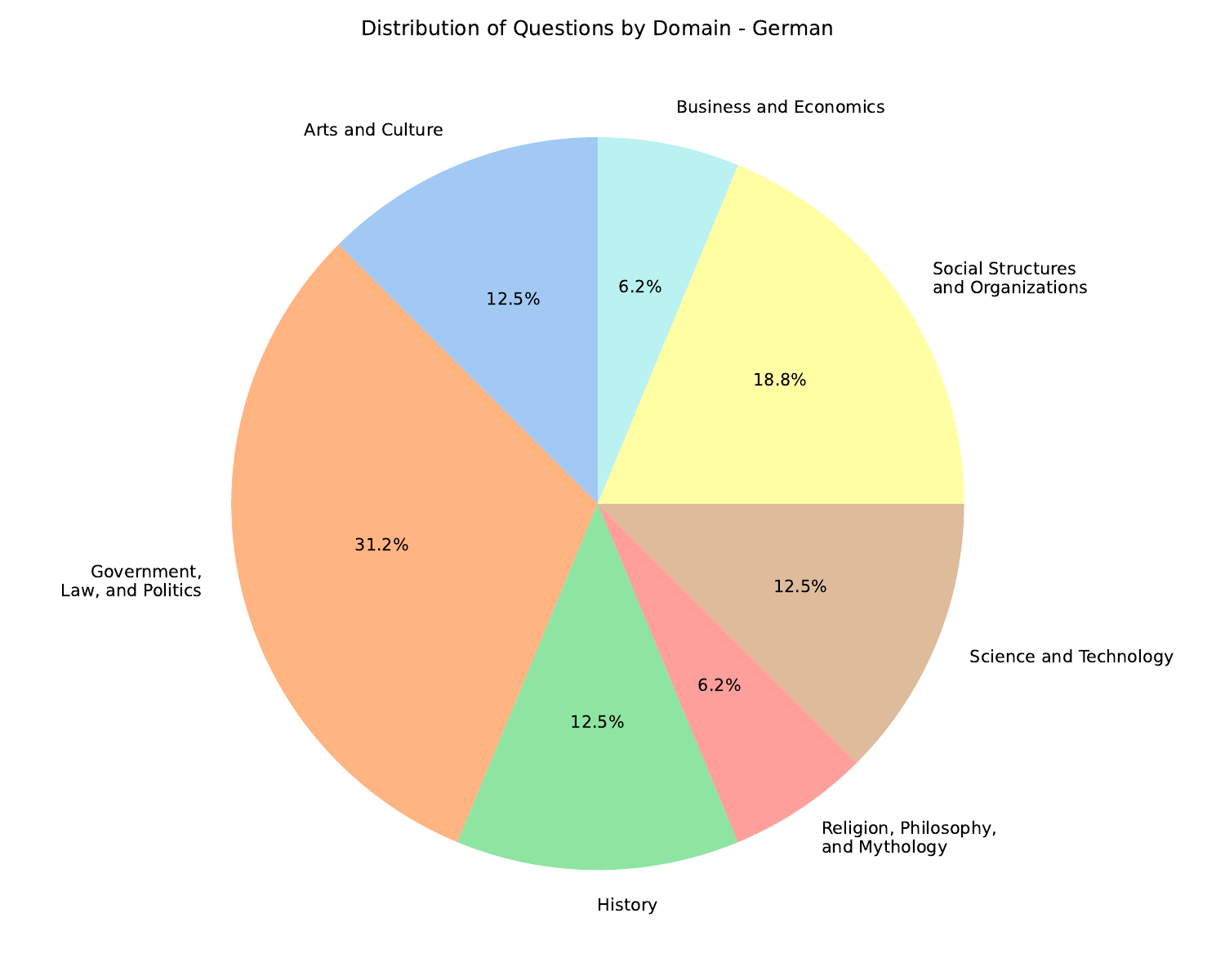}
        \label{fig:sub3}
    \end{subfigure}%
    \hfill
    \begin{subfigure}[b]{0.48\columnwidth}
        \includegraphics[width=\textwidth]{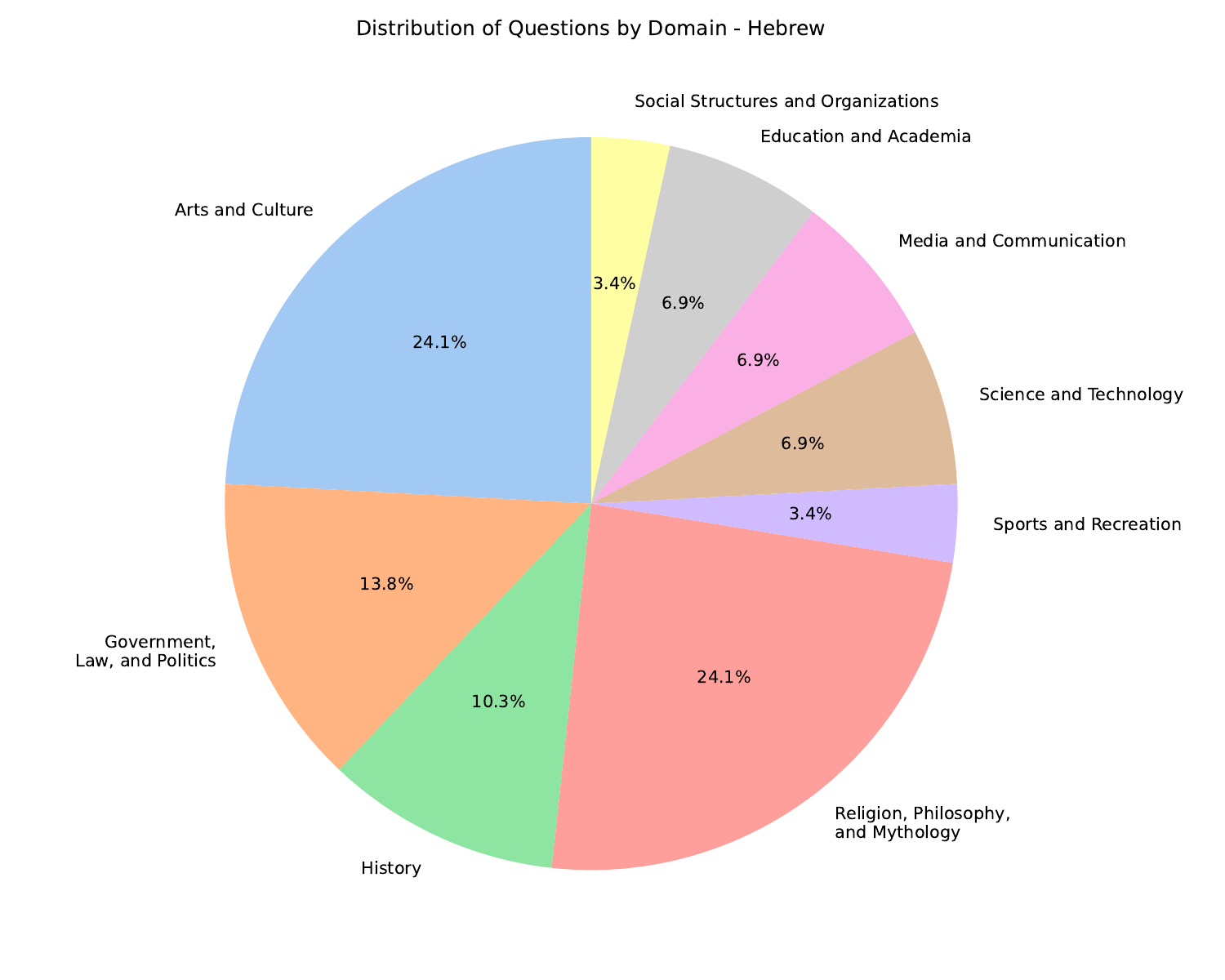}
        \label{fig:sub4}
    \end{subfigure}
    
    \vspace{\baselineskip}
    
    \begin{subfigure}[b]{0.48\columnwidth}
        \includegraphics[width=\textwidth]{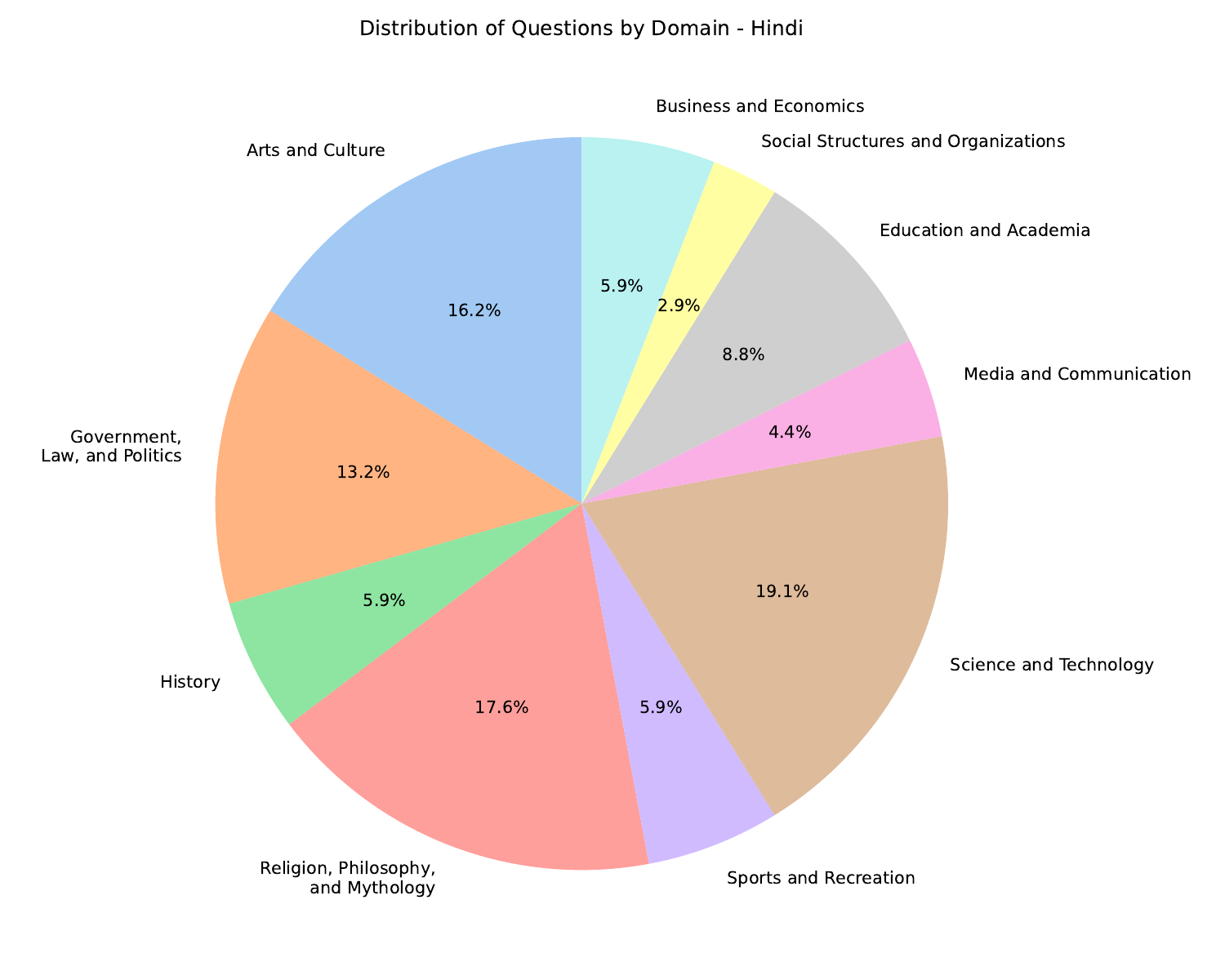}
        \label{fig:sub5}
    \end{subfigure}%
    \hfill
    \begin{subfigure}[b]{0.48\columnwidth}
        \includegraphics[width=\textwidth]{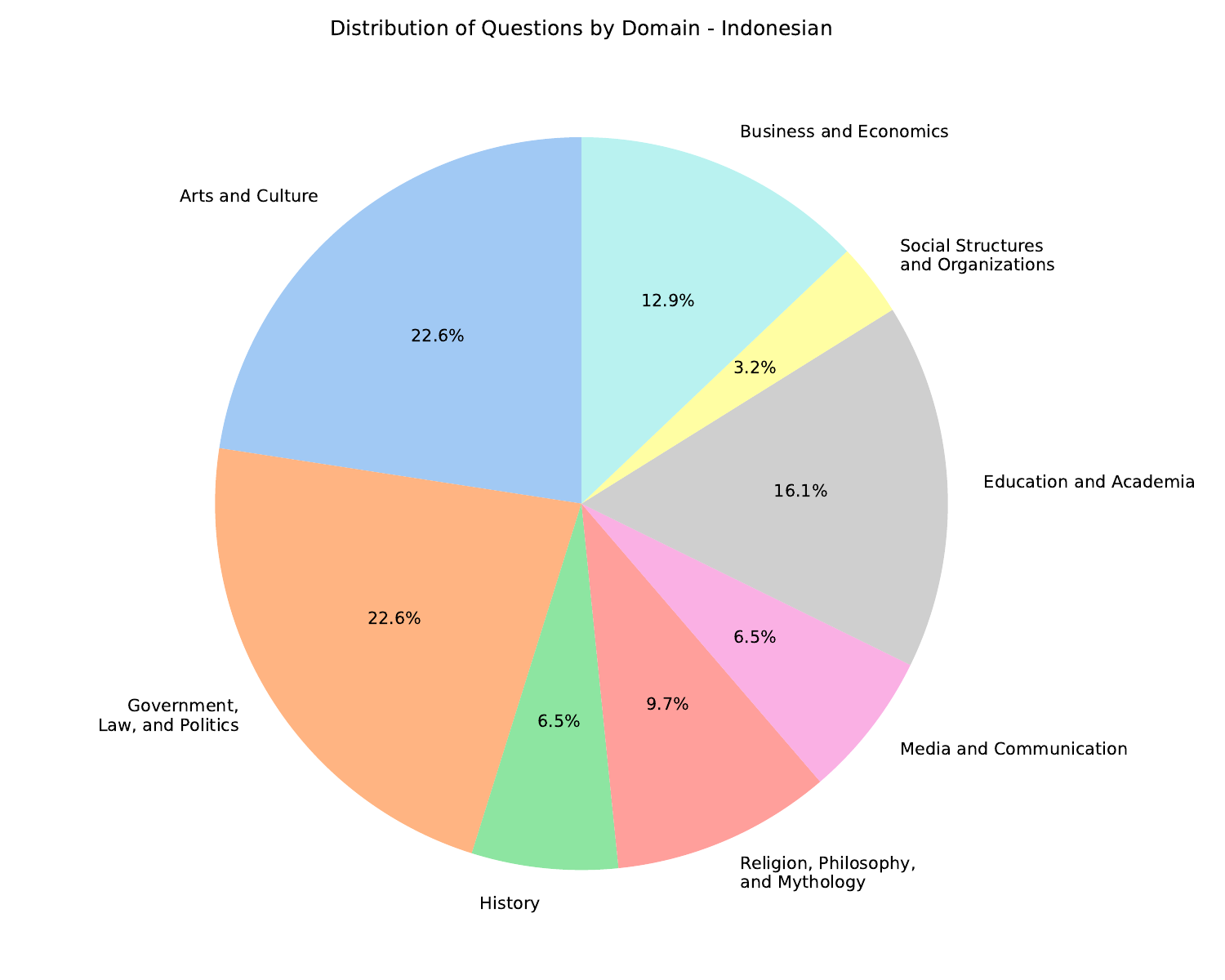}
        \label{fig:sub6}
    \end{subfigure}

    \caption{Distribution of the questions in \eclektic across domains. Per language breakdown.}
    \label{fig:all-breakdowns-a}
\end{figure*}

\begin{figure*}[htbp]

    \begin{subfigure}[b]{0.48\columnwidth}
        \includegraphics[width=\textwidth]{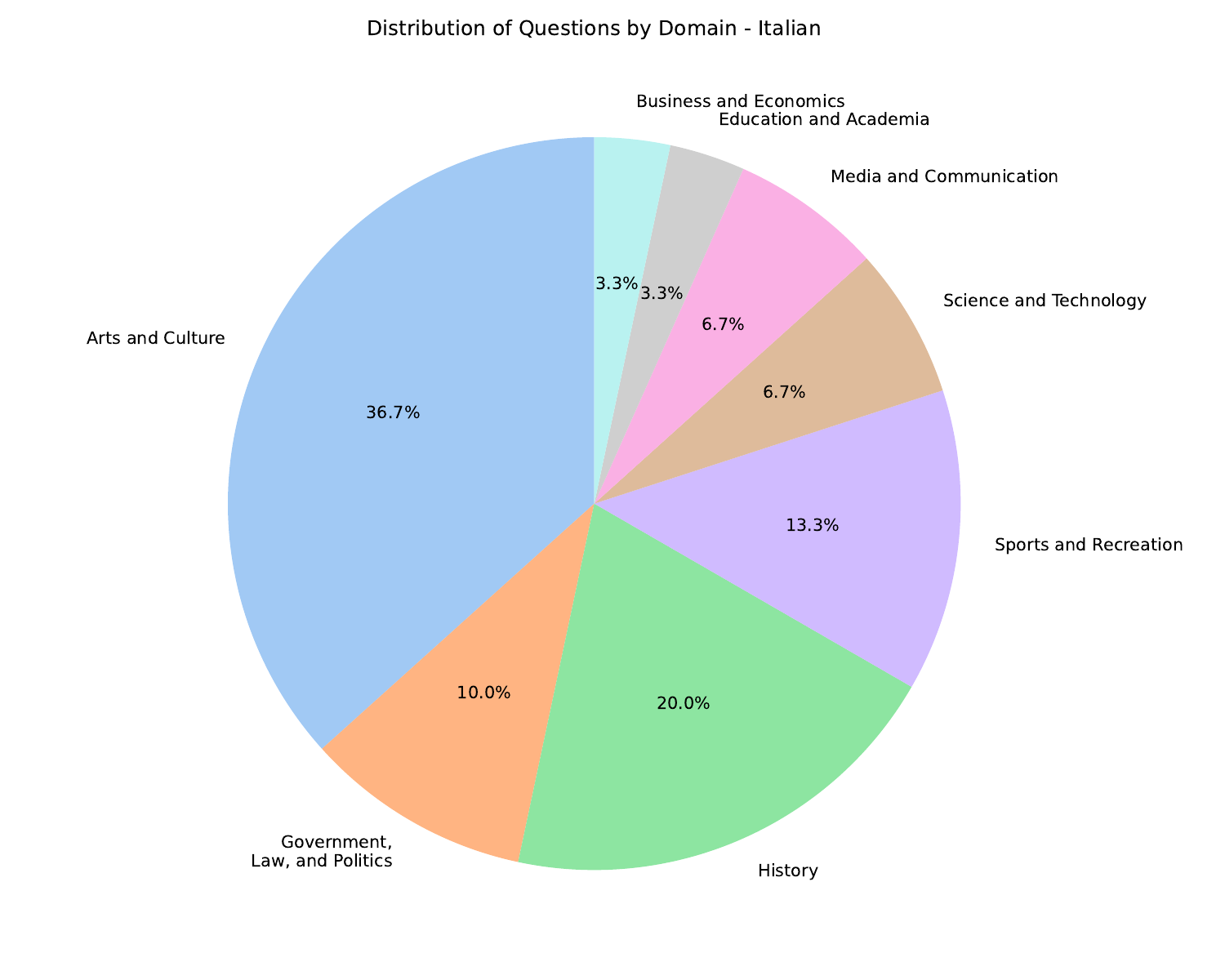}
        \label{fig:sub7}
    \end{subfigure}%
    \hfill
    \begin{subfigure}[b]{0.48\columnwidth}
        \includegraphics[width=\textwidth]{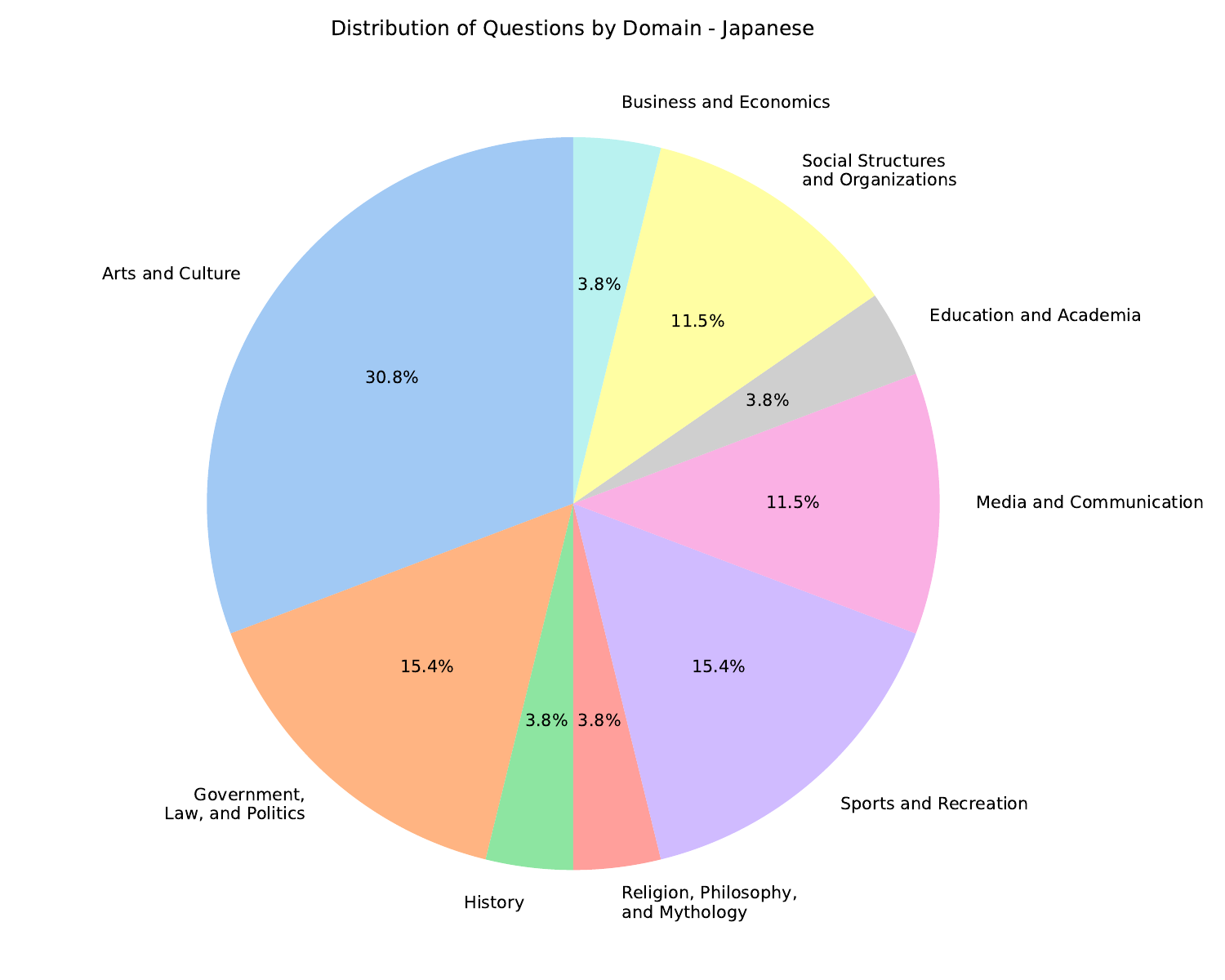}
        \label{fig:sub8}
    \end{subfigure}
    
    \vspace{\baselineskip}
    
    \begin{subfigure}[b]{0.48\columnwidth}
        \includegraphics[width=\textwidth]{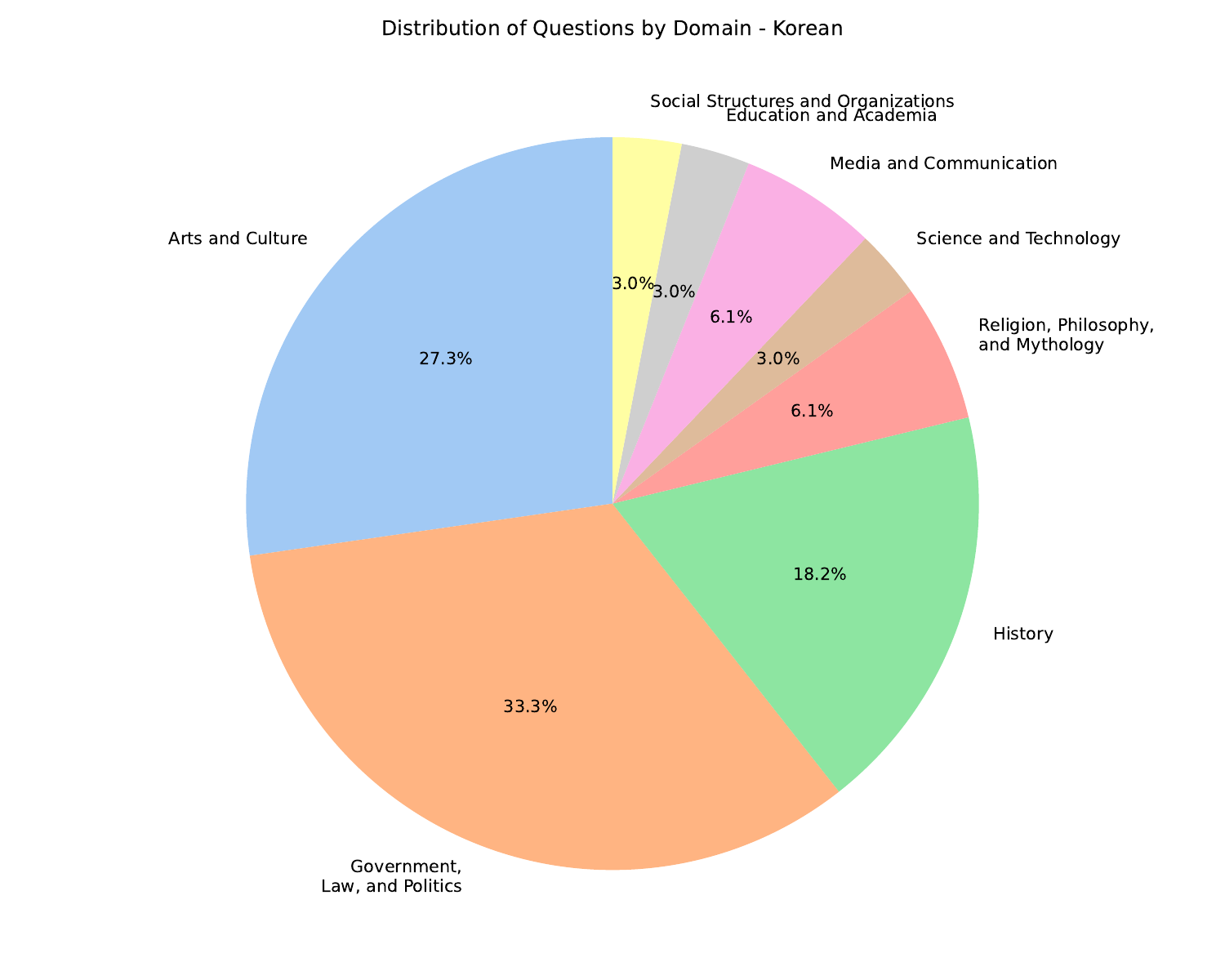}
        \label{fig:sub9}
    \end{subfigure}%
    \hfill
    \begin{subfigure}[b]{0.48\columnwidth}
        \includegraphics[width=\textwidth]{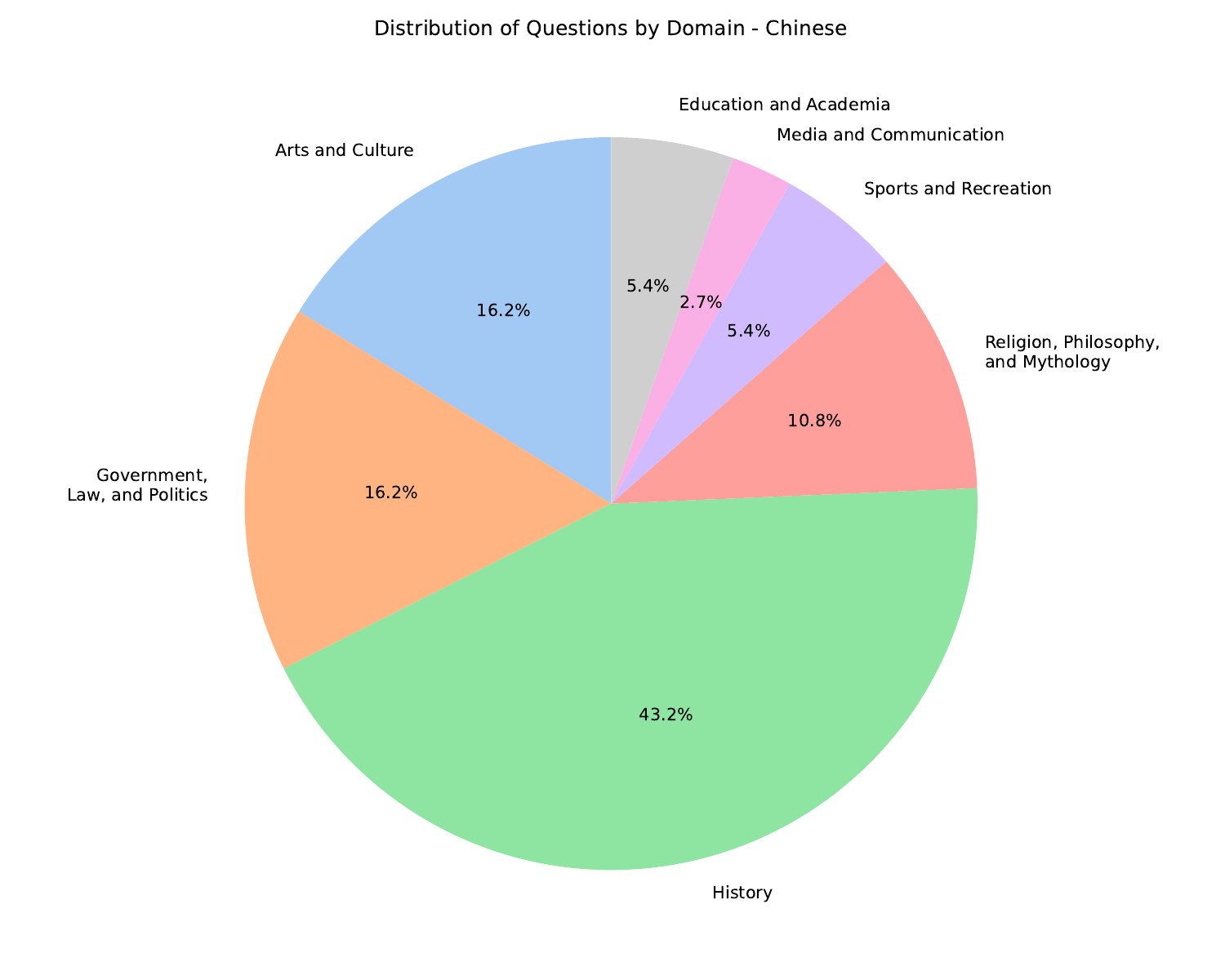}
        \label{fig:sub10}
    \end{subfigure}

    \vspace{\baselineskip}

    \begin{subfigure}[b]{0.48\columnwidth}
        \includegraphics[width=\textwidth]{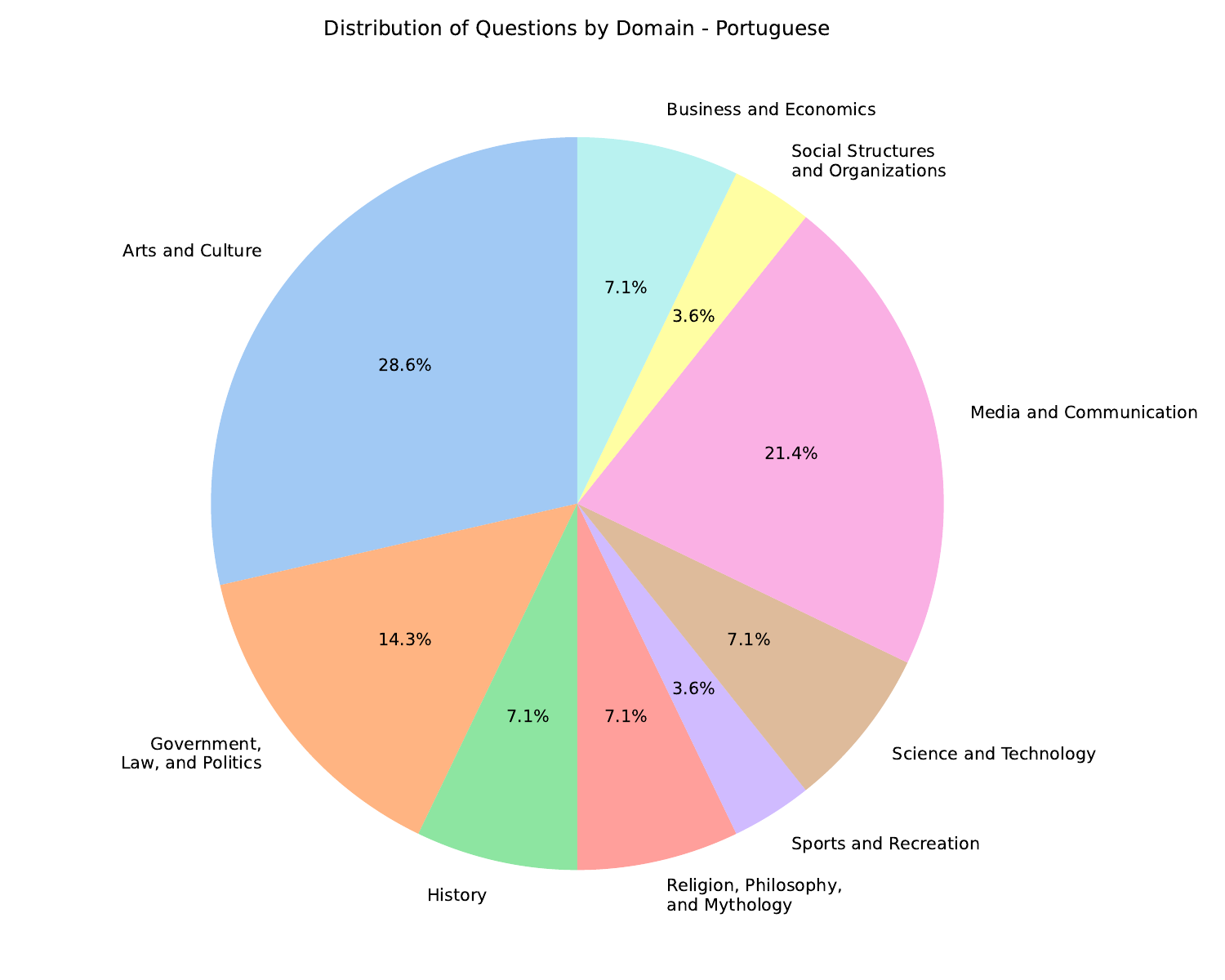}
        \label{fig:sub11}
    \end{subfigure}%
    \hfill
    \begin{subfigure}[b]{0.48\columnwidth}
        \includegraphics[width=\textwidth]{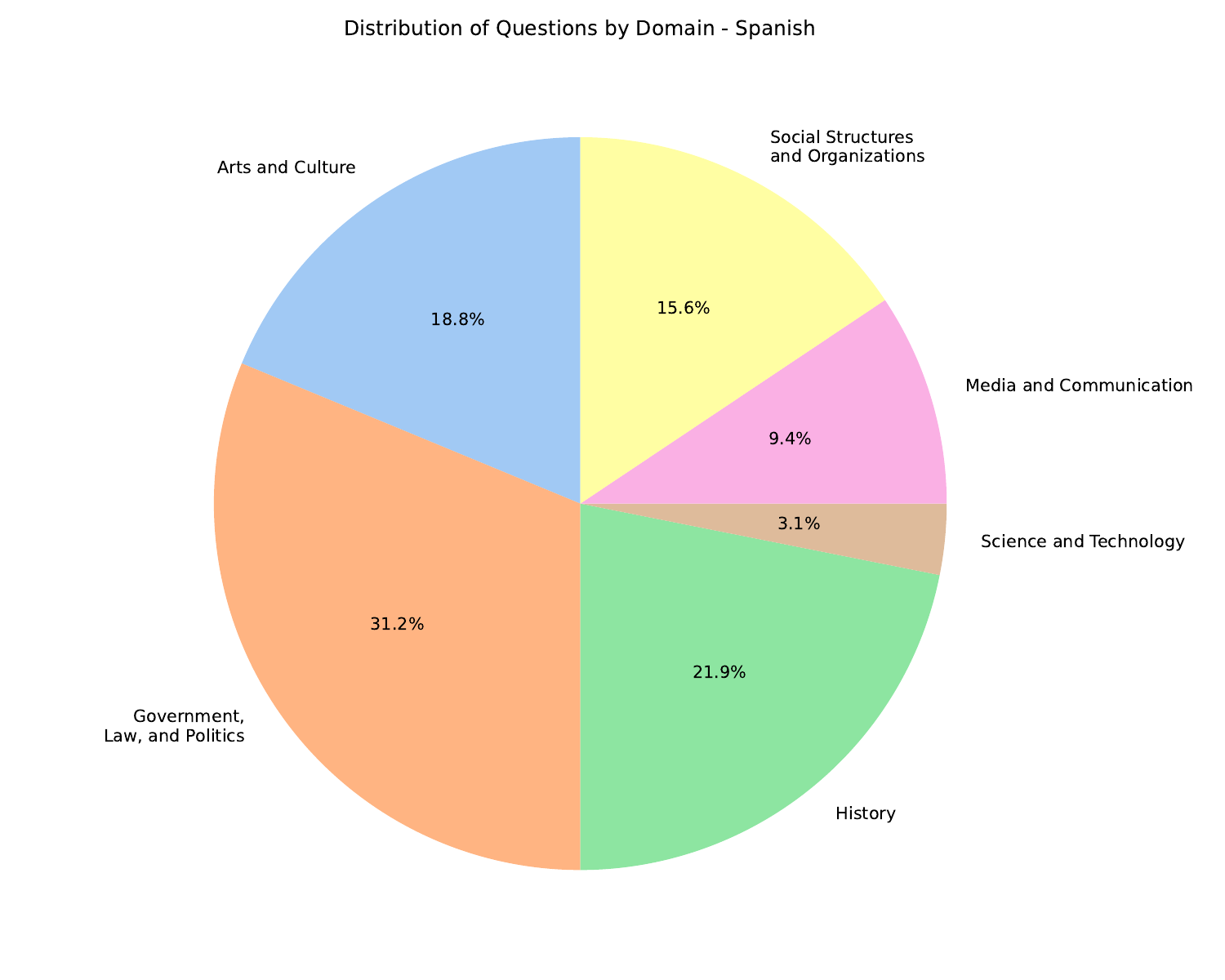}
        \label{fig:sub12}
    \end{subfigure}

    \caption{Distribution of the questions in \eclektic across domains. Per language breakdown -- continued.}
    \label{fig:all-breakdowns-b}
\end{figure*}

\hyperref[fig:all-breakdowns-a]{Figures}~\ref{fig:all-breakdowns-a} and~\ref{fig:all-breakdowns-b} show the break down per source language of questions in \eclektic by domain of knowledge.%

\end{document}